\documentclass[runningheads]{llncs}

\usepackage[mobile]{eccv}

\usepackage{eccvabbrv}

\usepackage{colortbl} 

\usepackage[accsupp]{axessibility}  

\usepackage[pagebackref,breaklinks,colorlinks,citecolor=eccvblue]{hyperref}

\usepackage{graphicx}
\usepackage{amsmath}
\usepackage{amssymb}
\usepackage{booktabs}
\usepackage{caption}
\usepackage{subcaption}
\usepackage{multirow}
\usepackage{wrapfig}
\usepackage{makecell}
\usepackage{diagbox} 
\usepackage{bm}
\definecolor{ForestGreen}{RGB}{34,139,34}
\definecolor{dino}{RGB}{249,231,227}
\definecolor{dino_text}{RGB}{209,154,128}

\usepackage{colortbl} 
\newcolumntype{b}{>{\columncolor{dino}}c}

\newlength\savewidth\newcommand\shline{\noalign{\global\savewidth\arrayrulewidth
		\global\arrayrulewidth 1pt}\hline\noalign{\global\arrayrulewidth\savewidth}}
\newcommand{\tablestyle}[2]{\setlength{\tabcolsep}{#1}\renewcommand{\arraystretch}{#2}\centering\footnotesize}
\renewcommand{\paragraph}[1]{\vspace{1.25mm}\noindent\textbf{#1}}

\newcolumntype{x}[1]{>{\centering\arraybackslash}p{#1pt}}
\newcolumntype{y}[1]{>{\raggedright\arraybackslash}p{#1pt}}
\newcolumntype{z}[1]{>{\raggedleft\arraybackslash}p{#1pt}}

\begin{document}
	
	\title{Attention-Challenging Multiple Instance Learning for Whole Slide Image Classification} 
	
	\titlerunning{ACMIL}
	
	\author{Yunlong Zhang\inst{1,2} \and
		Honglin Li\inst{1,2} \and
		Yunxuan Sun\inst{1,2} \and
		Sunyi Zheng\inst{2} \and
		Chenglu Zhu\inst{2} \and
		Lin Yang\inst{2}}
	
	\authorrunning{F.~Author et al.}
	
	\institute{Zhejiang University \and
		Westlake University\\}
	
	\maketitle

	\begin{abstract}
		In the application of Multiple Instance Learning (MIL) methods for Whole Slide Image (WSI) classification, attention mechanisms often focus on a subset of discriminative instances, which are closely linked to overfitting. To mitigate overfitting,  we present Attention-Challenging MIL (ACMIL). ACMIL combines two techniques based on separate analyses for attention value concentration. Firstly, UMAP of instance features reveals various patterns among discriminative instances, with existing attention mechanisms capturing only some of them. To remedy this, we introduce Multiple Branch Attention (MBA) to capture more discriminative instances using multiple attention branches. Secondly, the examination of the cumulative value of Top-K attention scores indicates that a tiny number of instances dominate the majority of attention. In response, we present Stochastic Top-K Instance Masking (STKIM), which masks out a portion of instances with Top-K attention values and allocates their attention values to the remaining instances. The extensive experimental results on three WSI datasets with two pre-trained backbones reveal that our ACMIL outperforms state-of-the-art methods. Additionally, through heatmap visualization and UMAP visualization, this paper extensively illustrates ACMIL's effectiveness in suppressing attention value concentration and overcoming the overfitting challenge.	The source code is available at \url{https://github.com/dazhangyu123/ACMIL}.
		\vspace{-1mm}
		\vspace{-1mm}
	\end{abstract}

	\section{Introduction}\vspace{-1mm}
	\label{sec:intro}

	Whole slide image (WSI) classification is a critical undertaking in digital pathology, aiming to extract valuable information from high-resolution scanned images for precise diagnosis \cite{he2012histology,li2018cancer,wang2016deep,wang2021predicting}, prognosis \cite{zhu2017wsisa,li2018graph,yao2020whole,chen2021multimodal}, and treatment planning \cite{cornish2012whole,litjens2016deep,madabhushi2009digital,pantanowitz2011review,pinckaers2020streaming} of diseases.
	In recent years, multiple instance learning (MIL) \cite{amores2013multiple,dietterich1997solving,maron1997framework} has emerged as a promising approach for WSI classification, treating each WSI as a "bag" and its extracted small patches as "instances" within the bag, thus enabling efficient classification of WSIs through assigning a single label to the entire slide.

	\begin{figure}[t]
		\begin{minipage}{0.48\textwidth}
			\centering
			\includegraphics[width=\linewidth]{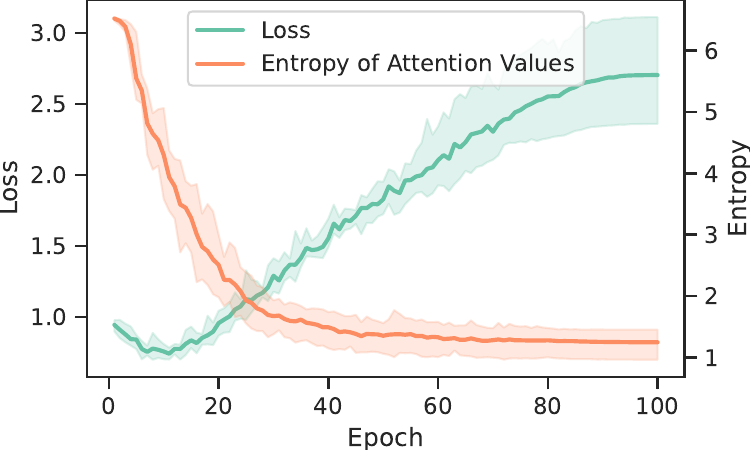}
			\captionof{figure}{The change of validation loss and entropy of attention values throughout the training of ABMIL. The results are reported on LBC with SSL pretrained features. There exists the strong negative correlation between loss and entropy. }
			\label{fig:loss_vs_entropy}
		\end{minipage}
		\hfill
		\begin{minipage}{0.48\textwidth}
			\centering
			\includegraphics[width=\linewidth]{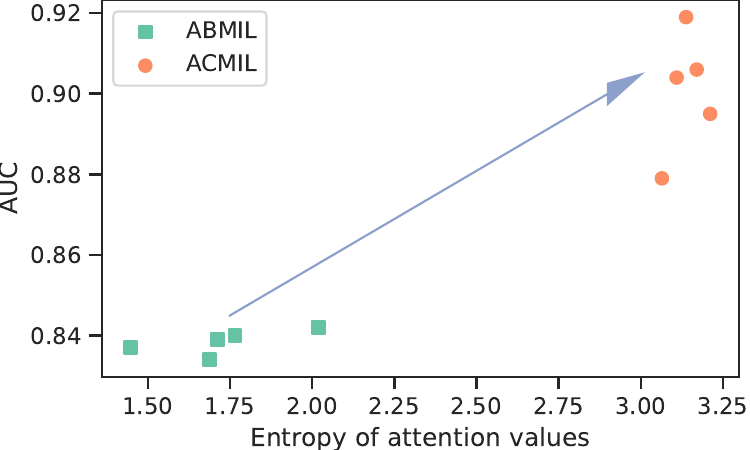}
			\captionof{figure}{ Comparison of AUC and entropy of attention values between ABMIL and ACMIL.  One point denotes the result of a seed on LBC with SSL pretrained features. ACMIL achieves the higher AUC and entropy than ABMIL.}
			\label{fig:auc_vs_entropy}
		\end{minipage}
		\vspace{-3mm}
	\end{figure}

	Overfitting is a significant challenge in utilizing MIL methods for WSI classification \cite{zhang2022dtfd,lin2023interventional,tang2023multiple}. Common WSI datasets exhibit intrinsic characteristics of limited data scale, ultra-high resolutions, and staining bias, which makes overfitting more likely \cite{bejani2021systematic}. Specifically, these datasets often consist of a relatively small number of slides, typically in the hundreds, with a resolution ranging from $50,000\times 50,000$ to $10,000\times10,000$ \cite{lu2021data}. Moreover,  pathology images are susceptible to staining bias caused by variations in tissue preparations, staining protocols, and digital scanning methods \cite{zhang2022benchmarking}, leading models to learn spurious correlations \cite{lin2023interventional}.

	In the attention mechanism, attention values/heatmaps provide insights into the model's decision-making process. Multiple existing works \cite{lu2021data,shao2021transmil,tang2023multiple,yufei2022bayes} alongside our own experiments (as indicated in Sec. \ref{sec:localization}) have pointed out the excessive concentration of attention values in current MIL methods. Specifically, MIL's attention mechanisms often concentrate on a subset of discriminative instances (i.e., instances relevant to the bag label) while disregarding the remaining ones. We investigate the correlation between attention value concentration and overfitting, utilizing the entropy of attention values and validation loss.  Fig. \ref{fig:loss_vs_entropy} depicts a negative correlation between loss and entropy throughout the training process, illustrating that over-concentration of attention values (indicated by lower entropy) significantly compromises the model's generalization ability (indicated by higher loss values). Moreover, in the field of natural image classification, recent studies \cite{huang2020self,tiwari2023overcoming,devries2017improved} have demonstrated that models solely relying on a portion of discriminative features could be susceptible to overfitting. Transitioning to WSI classification, fixating on a subset of discriminative instances similarly impedes the model's ability to generalize.  These findings highlight the tight connection between attention value concentration and overfitting. 
	
	Recently, numerous efforts have been made to address the overfitting challenge by enhancing representation quality \cite{lu2021data, dehaene2020self, holdenried2023dual, wang2022scl,li2023task}, building spatial instance correlations \cite{li2021dual,shao2021transmil,guan2022node} and developing data augmentation methods \cite{zhang2022dtfd,chikontwe2020multiple,yang2022remix,tang2023multiple,qu2022bi}. Additionally, some of these studies \cite{lu2021data,li2021dual}  suggest that reducing the concentration of attention values can enhance model interpretability. However, the investigation of attention values concentration for alleviating overfitting remains under-explored.

	To mitigate overfitting, we present two analyses for attention value concentration using UMAP and Top-K value statistics.
	Then, we introduce Attention-Challenging MIL (ACMIL), which combines two novel techniques based on these two analyses. First, by observing UMAP of instance features, we find that \textit{there are various patterns among discriminative instances, and attention mechanisms tend to capture some of them.} To solve this, we introduce Multiple Branch Attention (MBA). MBA utilizes multiple attention branches, each focusing on capturing instances with a specific pattern, thereby ensuring that more discriminative instances contribute to the final prediction. Second, by analyzing the cumulative value of Top-K attention scores, we find that \textit{ a tiny number of instances (e.g., K=10) occupy majority attention}, resulting in overlooking sophisticated discriminative instances. To suppress these instances, we propose Stochastic Top-K Instance Masking (STKIM). STKIM randomly masks out a portion of instances with Top-K attention values and assigns their attention values to the remaining instances. Combining MBA and STKIM, our ACMIL effectively alleviates the attention value concentration and suppresses overfitting (see Fig. \ref{fig:auc_vs_entropy}).

	We conduct experiments on three WSI datasets (i.e., CAMELYON16, BRACS, and our in-house LBC dataset) with two backbones (i.e., ImageNet pre-trained ResNet18 and SSL pre-trained ViT/S-16). Experimental results demonstrate the superiority of our ACMIL over existing state-of-the-art methods. We also present substantial experimental results, including heatmap visualization and UMAP visualization, to comprehensively demonstrate the effectiveness of ACMIL in suppressing attention value concentration and combatting overfitting.

	\section{Related Work}\vspace{-1mm}
	\subsection{Combating Overfitting in WSI Analysis}\vspace{-1mm}

	In the domain of WSI classification, combating the challenge of overfitting has received substantial attention. Next three paragraphs detail methods from three different aspects.
	
	Some efforts have concentrated on enhancing the quality of instance representations. Early studies (e.g., \cite{ilse2018attention, campanella2019clinical, shao2021transmil}) rely on backbones pre-trained on the ImageNet dataset. However, the substantial domain gap between natural and pathological images hinders representation quality. Recent works (e.g., \cite{tellez2019neural,lu2021data, dehaene2020self, holdenried2023dual, wang2022scl}) address this by emphasizing Self-Supervised Learning (SSL) to learn patch-level feature representations. In addition, efforts such as the work by Chen et al. \cite{chen2022scaling} leverage hierarchical SSL for high-resolution image representations. Further, studies by Li et al. \cite{li2023task} and Wang et al. \cite{wang2023iteratively} demonstrate that fine-tuning the pre-trained encoder is essential for acquiring task-specific information.
	
	Another line of research has focused on establishing spatial instance correlations. DSMIL \cite{li2021dual}, H$^2$MIL \cite{hou2022h}, and DAS-MIL \cite{bontempo2023graph} consider the hierarchical structure of patches and aggregate multi-scale representations in attention mechanisms. Furthermore, some studies introduce self-attention layers \cite{shao2021transmil, xiong2023diagnose, qu2023boosting} and graph neural networks \cite{guan2022node, zhao2020predicting,chan2023histopathology} to model correlations between different areas.
	
	Further strategies have concentrated on data augmentation. Examples include DTFD-MIL \cite{zhang2022dtfd}, which introduces pseudo-bags for expanding bag counts and employs a double-tier MIL framework. IPS \cite{bergner2022iterative}, Zoom-In Network \cite{kong2022efficient}, and Top-K MIL \cite{chikontwe2020multiple} generate bag representations by aggregating the representations of salient patches. Remix \cite{yang2022remix} and RankMix \cite{chen2023rankmix} introduce instance representation mixup for MIL. MHIM-MIL \cite{tang2023multiple} and WENO \cite{qu2022bi} augments bags by randomly masking salient instances.
	
	Although our ACMIL shares a similar spirit with some of these works, the proposed ACMIL excels by further building on detailed analysis for attention value concentration. As a result, our ACMIL exhibits stronger interpretability against existing solutions.

	\subsection{Over-Concentration of Attention Values}\vspace{-1mm}
	In the realm of natural image classification, research has shown that an excessive focus on certain parts of an object can impede the overall effectiveness of model generalization \cite{zhong2020random,devries2017improved,huang2020self}. To tackle this issue, various heuristic techniques have been proposed. For instance, Cutout \cite{zhong2020random,devries2017improved} is a valuable data augmentation method that randomly masks square regions of input during training. Another approach, RSC \cite{huang2020self}, involves regularization that eliminates salient features activated during training. This paper investigates the issue of attention value concentration in WSI classification tasks. We identify two specific phenomena related to attention value concentration existing in WSI classification and propose two techniques to address them respectively.

	\section{Method}\vspace{-1mm}
	
	Based on the ABMIL (detailed in Sec. \ref{sec31}), we present ACMIL to alleviate the overfitting problem, which is built on two components: Multiple Branch Attention (MBA) and Stochastic Top-K Instance  Masking (STKIM). We describe the details of two components in Sec. \ref{sec32} and \ref{sec33}, respectively.
	\subsection{ABMIL for WSI Classification} \label{sec31}
	In the binary MIL classification problem \cite{campanella2019clinical}, a bag of instance, $\bm X = \{\bm x_n\}_{n=1}^N$, is associated with a single bag label, $\bm Y$.  Each instance, $x_n$, is associated with a single binary label, $y_n$, which remains unknown during training. The assumption behind the MIL can be written as:
	\begin{equation}
		\bm Y= \begin{cases}0, & \text { iff } \sum_{n=1}^{N} y_n=0 \\ 1, & \text { otherwise } \end{cases}
	\end{equation}

	In the ABMIL \cite{ilse2018attention}, the multiple instance learning is modeled by a three-step process. i) Instance transformation into a low-dimensional embedding through neural networks: $\bm h_n = f(\bm x_n)$. ii) Aggregation of all instance embeddings into the bag-level representation using an attention operator. Specifically, this operation is defined as:
	\begin{equation} \label{eq2}
		\bm{z} = \sum_{n=1}^{N} a_n \bm{h}_{n}
	\end{equation}
	Here, $a_n = \sigma (\bm{h}_{n})$ represents the attention values for $n$-th instance, $\bm{h}_{n}$. In the case of ABMIL, a gated attention (GA) mechanism \cite{dauphin2017language} is adopted:
	\begin{equation} \label{eq3}
		\sigma (\bm{h}_{n}) = \frac{ \exp \{ \bm{w}^{\text{T}} ( \text{tanh}(\bm{V}_1 \bm{h}_n ) \odot \text{sigm}( \bm{V}_2 \bm{h}_n ) )  \}  }{  \sum_{j=1}^{N}  \exp \{ \bm{w}^{\text{T}} ( \text{tanh}(\bm{V}_1 \bm{h}_j ) \odot \text{sigm}( \bm{V}_2 \bm{h}_j ) )  \}   }
	\end{equation}
	where $\bm{V}_1, \bm{V}_2 \in \mathbb{R}^{L \times M}$, $\bm{w} \in \mathbb{R}^{L \times 1}$ are parameters, $\odot$ is an element-wise multiplication and $\mathrm{sigm(\cdot)}$ is the sigmoid non-linearity.
	iii) The bag prediction is generated based on the aggregated bag embedding: $\hat{\bm Y} = g(\bm z)$.

	\subsection{Mutiple Branch Attention} \label{sec32}
	\begin{wrapfigure}{r}{0.5\textwidth}
		\vspace{-8mm}
		\centering
		\includegraphics[width=0.45\textwidth]{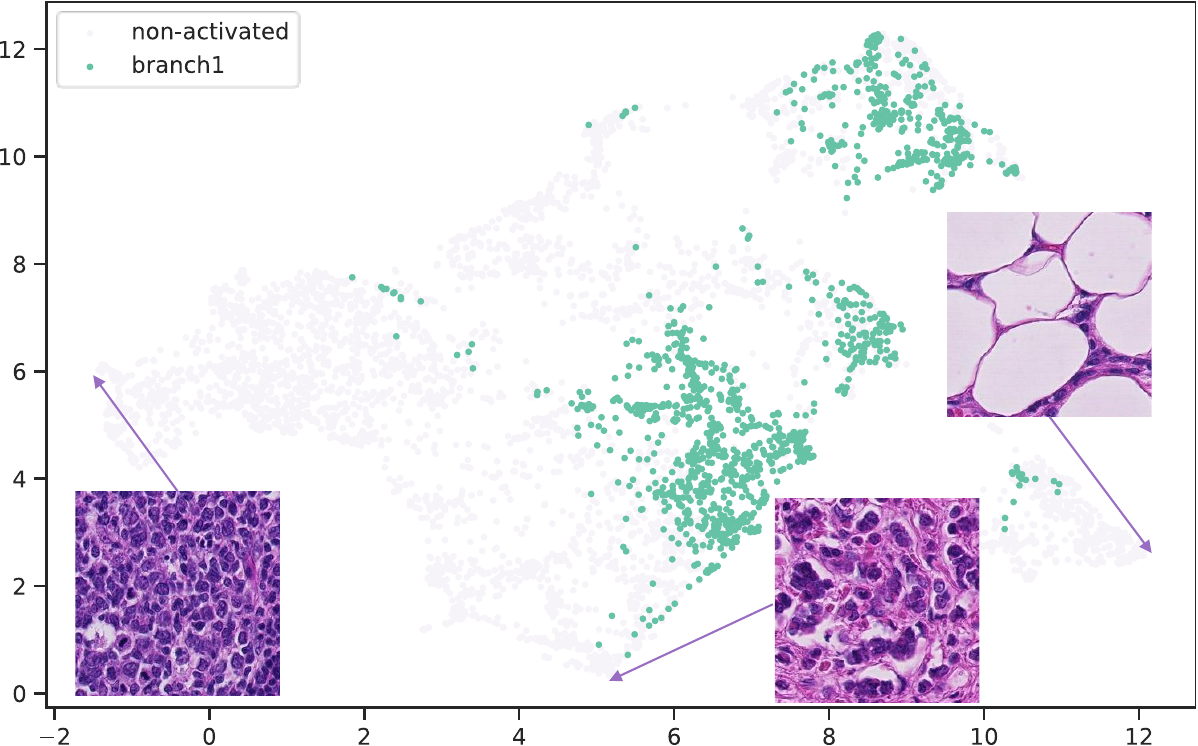}
		\caption{Motivation of MBA. UMAP visualization \cite{mcinnes2018umap} of tumor instance features from CAMELYON16 `test\_113' case. \textit{There are various patterns/clusters among tumor instances, and relying on one single branch tends to capture a part of clusters.} Three instances are selected to exhibit their texture differences.  }
		\vspace{-8mm}
		\label{fig_mba}
	\end{wrapfigure}
	\textbf{Motivation.} It is challenging to capture all discriminative instances using a single attention branch (see Fig. \ref{fig_mba}). This challenge arises due to variations in patterns among discriminative patches, stemming from differences in texture and morphology. Additionally, DNNs tend to exhibit a form of "laziness" where they prioritize capturing simpler patterns to minimize training loss, neglecting more intricate and challenging patterns \cite{geirhos2018imagenet,geirhos2020shortcut}. To tackle this issue, we design the MBA that captures more discriminative instances by multiple attention branches.

\begin{figure*}[t]
		\centering
		\includegraphics[width=0.95\textwidth]{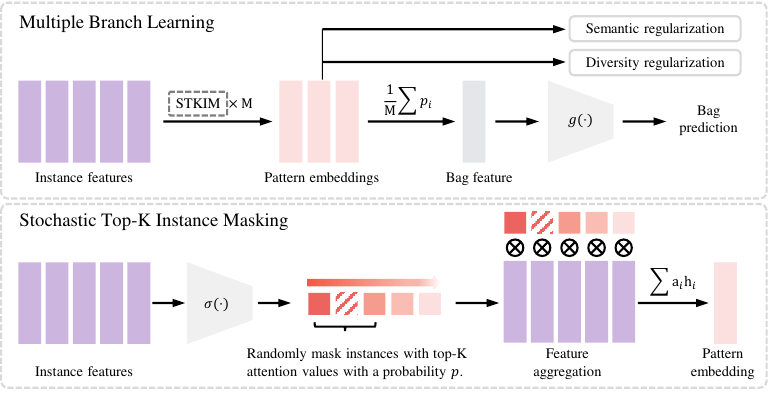}
		\caption{Overview of the proposed MBA (top view) and STKIM (bottom view). In the MBA, $M$ discriminative patterns are extracted from patch features using the attention operator regularized by semantic and diversity regularization terms. Then, the mean operator is applied to these $M$ pattern features to produce the bag feature, which is utilized for bag-level prediction. In the STKIM,  instances with Top-K attention values are randomly masked  with a probability $p$.}
		\label{fig_overview}
	\end{figure*}

	As depicted in Fig. \ref{fig_overview} top view, the MBA firstly captures $M$ patterns and then aggregates their embeddings to make predictions. Each pattern is captured by an attention branch. To maintain both the discriminative nature of patterns and semantic diversity between them, we introduce two regularization techniques: semantic regularization and diversity regularization. Firstly, to ensure capturing discriminative patterns, the semantic regularization is accomplished by hanging a MLP layer behind each pattern embedding, equipping with the cross entropy loss function:
	\begin{equation} \label{eq4}
		\mathcal{L}_p =  - \frac{1}{M} \sum_{i=1}^{M}  \bm Y \log{\bm \hat{\bm Y}_i} + \left(1-\bm Y \right) \log{ \left( 1-\bm \hat{\bm Y}_i \right) }
	\end{equation}
	where $\bm \hat{\bm Y}_i = g_i(\bm z_i)$ is the prediction based on $i$-th pattern embedding, $\bm z_i$.  However, only equipping with cross-entropy loss may learn similar patterns and cannot dig out more discriminative information. To tackle this issue, we further introduce a diversity loss as follows: 
	\begin{equation} \label{eq5}
		\mathcal{L}_d = \frac{2}{M(M-1)} \sum_{i=1}^{M} \sum_{j=i+1}^{M} \cos(\bm a_i, \bm a_j)
	\end{equation}
	where $\bm a_i$ consists of all attention values of $i$-th pattern, $\bm a_i = \{a_{i1}, \cdots, a_{iN}\}$, also named heatmap as custom. The $\cos(\cdot)$ function is used to measure the similarity of the heatmaps between branches. By diversifying the heatmaps, the embedding of each branch can concentrate on different patterns.
	

	
	To aggregate the captured patterns to make predictions, the average of heatmaps is utilized as the heatmap of the whole bag:
	\begin{equation} \label{eq6}
		\bm a = \frac{1}{M}\sum_{i=1}^M\bm a_i
	\end{equation}
	where $\bm a$ is the heatmap of the whole bag, with a dimension of $N$.  Then,  the bag embedding can be obtained by aggregating the instance features using averaged heatmap $\bm a$. Moreover, since $\sum_{n=1}^{N}(\frac{1}{M}\sum_{i=1}^M a_{in}) \bm{h}_{n} = \frac{1}{M}\sum_{i=1}^M (\sum_{n=1}^{N} a_{in} \bm{h}_{n})$, the bag embedding also can be formulated by applying mean pooling operator to pattern embeddings. The top view of Fig. \ref{fig_overview} adopts the latter formulation for brevity. The loss function for the bag classifier is defined as:
	\begin{equation} \label{eq_lb}
		\mathcal{L}_b =  -   \bm Y \log{\bm \hat{\bm Y}} + \left(1-\bm Y \right) \log{ \left( 1-\bm \hat{\bm Y} \right) }
	\end{equation}
	Finally, the overall loss function for the ACMIL can be written as the combination of three loss terms defined in Eq. \ref{eq4}, \ref{eq5} and \ref{eq_lb},
	\begin{equation} \label{eq321}
		\mathcal{L} =  \mathcal{L}_b + \mathcal{L}_p + \mathcal{L}_d
	\end{equation}

	\noindent\textbf{Discussion.} It's important to highlight that when the parameter $M$ is set to 1 in MBA, it essentially mirrors the feature aggregation process of ABMIL, allowing for the discernment of a single pattern. In this sense, MBA serves as an extension of ABMIL, specifically designed to capture a more diverse set of patterns. We further discuss the connection between MBA and Multiple-Head Attention (MHA). 	 HIPT \cite{chen2022scaling} has unveiled that distinct heads in MHA can effectively capture different visual concepts, akin to the role played by our MBA. However, these two techniques can be easily distinguished by: 1) MBA has diversity regularization, ensuring that different branches can learn different concepts. This is absent in MHA, resulting in different heads learning the same concept \cite{chen2022scaling}.  We demonstrate the critical role of $\mathcal{L}_d$ for performance in Tab. \ref{tab:T-sTKIM_ld_b}. 2) MHA is a type of attention formulation, while MBA operates independently of the attention formulation, accommodating MHA within its framework. Appendix Sec. \ref{sec_base} reports the results of combining MHA and ACMIL.
	
	%

	\subsection{Stochastic Top-K Instance Masking}\label{sec33}
	\begin{wrapfigure}{r}{0.5\textwidth}
		\vspace{-8mm}
		\centering
		\includegraphics[width=0.45\textwidth]{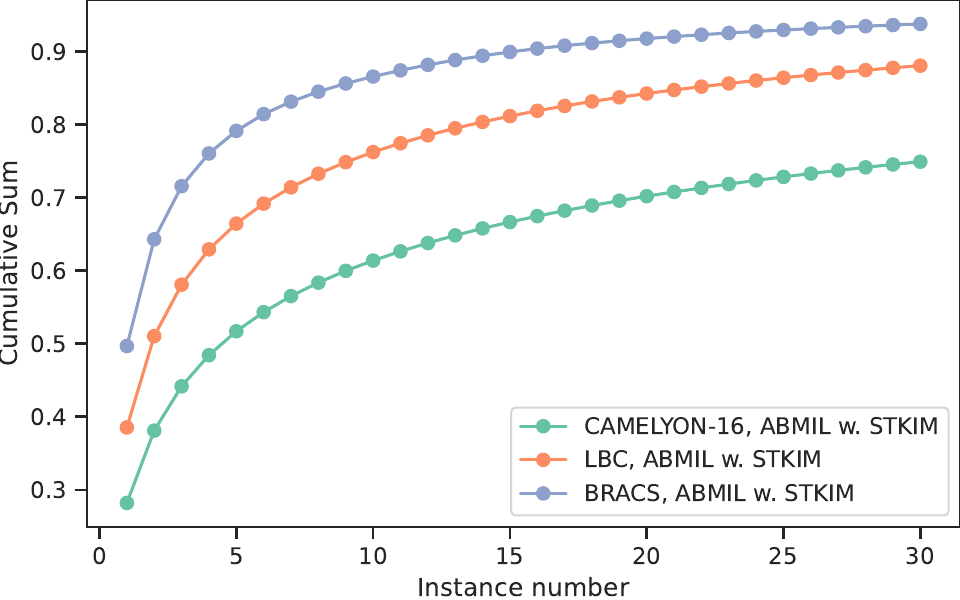}
		\caption{Motivation of STKIM. Accumulation of Top-K attention values. \textit{Instances with Top-K attention values occupy majority attention.} Results are derived from features extracted through supervised pretraining. }
		\vspace{-4mm}
		\label{fig_stkim}
	\end{wrapfigure}
	
	\textbf{Motivation.} A tiny number of instances will occupy the majority of attention in ABMIL while ignoring sophisticated discriminative instances. As depicted in Fig. \ref{fig_stkim}, the sum of top-10 attention values is larger than 0.85 on all three datasets.  However, the WSI typically involves more than 10 discriminative instances. For instance, in the CAMELYON16 dataset, 129 out of 155 tumor slides contain 10 to 20,000 cancerous instances. In essence, numerous discriminative instances are overlooked. To deal with this issue, the proposed STKIM aims to suppress the salient instances and assign more attention to the remaining instances.

	As depicted in Fig. \ref{fig_overview} bottom view, STKIM introduces a masking operation into the attention mechanism, before feature aggregation and after attention values generation. The primary objective is to suppress Top-K salient instances. A straightforward solution to achieve this is to mask out all of the Top-K salient instances. However, this method poses certain challenges. It can result in the loss of information associated with key instances, which are crucial for discrimination. Furthermore, it might lead to a statistical mismatch between the feature representations before and after discarding these key instances. To address these issues, we draw inspiration from dropout \cite{srivastava2014dropout} and cutout \cite{devries2017improved,zhong2020random} commonly used in computer vision. Our proposed solution employs stochastic masking for instance features with Top-K attention values.
	Specifically, we begin by sorting all attention values from highest to lowest. Subsequently, we randomly set the attention values of the Top-K instances to 0, with a probability of $p$. This process can be formulated as:
	\begin{equation} \label{eq8}
		a_n = \begin{cases}
			0, & \text{with probability } p \text{ and within Top-K values} \\
			a_n, & \text{otherwise}
		\end{cases}
	\end{equation}
	where $p$ and $K$ are two hyperparameters that control the intensity of masking. Following Eq. \ref{eq8}, we will assign the attention values of masked instances to the remaining instances by $a_n \rightarrow \frac{1}{\sum_{n=1}^{N}a_n}a_n$. Notably, drawing inspiration from dropout and cutout, we remove STKIM  at the inference time.

	\label{tech} 
	\noindent\textbf{Discussion.} While STKIM, MHIM-MIL \cite{tang2023multiple}, and WENO \cite{qu2022bi} all adopt the technique that masks salient instances, there are notable technical distinctions between them. Firstly, our STKIM has the different masking strategy compared with WENO and MHIM-MIL. STKIM only masks a minority of instances (i.e., $K=10$) with a probability of $p$. As a comparison, the other two methods mask out a larger number of instances. WENO masks out 95 instances. MHIM-MIL masks  $1\%$ instances. In our framework, our scheme performs best in three strategies (see Appendix Sec. \ref{sec_mask}). Secondly, both MHIM-MIL and WENO necessitate a well-trained model for masking out salient instances, utilizing the remaining instances for model training. They both employ a teacher-student framework, wherein the teacher model needs to be pre-trained beforehand (the warm-up process in WENO and the pre-training stage in MHIM-MIL). In contrast, STKIM requires neither a teacher-student framework nor a pre-training process, thus highlighting simplicity and efficiency.

	\section{Experiments}
	\subsection{Experimental Details}
	\noindent\textbf{Datasets and Evaluation Metrics.} The performance of ACMIL is evaluated on two public WSI datasets, i.e., CAMELYON16 \cite{bejnordi2017diagnostic} and BRACS \cite{brancati2022bracs}, and one private benchmark, LBC. CAMELYON16 dataset consists of 400 WSIs in total, including 270 for training and 130 for testing. Following \cite{zhang2022dtfd,li2021dual}, we further randomly split the training and validation sets from the official training set with a ratio of 9:1. We do not resplit BRACS dataset as it has been officially split to 395 of training set, 65 of validation set, and 87 of test set. We follow the challenge for a 3-class WSI classification: benign tumor, atypical tumor, and malignant. The liquid-based cytology (LBC) dataset collected 1,989 WSIs and included 4 classes, i.e., Negative, ASC-US, LSIL, and ASC-H/HSIL. We randomly split the whole dataset into training, validation, and test sets with the ratio of 6:2:2.  Following \cite{li2023task}, macro-AUC and macro-F1 scores are reported since all three datasets are class imbalanced. Each of the main experiments is performed five times with random parameter initializations, and the average classification performance and standard deviation are reported. Besides, following \cite{lu2021data,zhang2022dtfd}, the test performance is reported in epochs with the best validation performance.

	\noindent\textbf{Baselines.} We systematically assess the efficacy of our ACMIL approach by benchmarking it against conventional MIL pooling strategies, Max-pooling and Mean-pooling, as well as contemporary attention-based techniques such as ABMIL \cite{ilse2018attention}, DSMIL \cite{li2021dual}, TransMIL \cite{shao2021transmil}, CLAM-SB \cite{lu2021data}, DTFD-MIL \cite{zhang2022dtfd}, MHIM-MIL \cite{tang2023multiple}, and IBMIL \cite{lin2023interventional}. In pursuit of a comprehensive comparison across diverse aggregation operators, we utilize two distinct sets of features derived from ResNet-18 pre-trained on the ImageNet dataset \cite{he2016deep} and ViT-S/16 pretrained using DINO \cite{caron2021emerging} on a substantial collection of 36,666 WSIs \cite{kang2023benchmarking}. The results of all other methods are reproduced using the official code they provide under
	the same settings.
	
	\noindent\textbf{Implementation Details.} Implementation Details are in Appendix Sec. \ref{sec_imple}.

	\begin{table*}[t]
		\scriptsize
		\centering
		\caption{The performance of different MIL approaches across three datasets, two pre-trained methods, and two evaluation metrics. The most superior performance is highlighted in \textbf{bold}, while the second-best performance is indicated by \underline{underlining}. }
		\vspace{-1mm}
		\begin{tabular}{clcccccc}
			\toprule
			
			\multicolumn{2}{c}{\multirow{2}{*}{\diagbox[width=8em]{{Method}}{{Performance}}}} & \multicolumn{2}{c}{CAMELYON-16} & \multicolumn{2}{c}{BRACS} &  \multicolumn{2}{c}{LBC} \\ 
			\cmidrule(lr){3-4} \cmidrule(lr){5-6} \cmidrule(lr){7-8}
			&& F1-score       & AUC           & F1-score       & AUC       & F1-score            & AUC   \\\midrule
			\multirow{9}{*}{\rotatebox{90}{\makecell{ResNet-18 \\ImageNet pretrained}}} &
			Max-pooling & 0.582\tiny{$\pm$0.170} & 0.620\tiny{$\pm$0.155} & 0.489\tiny{$\pm$0.047} & 0.738\tiny{$\pm$0.014} & 0.476\tiny{$\pm$0.033} & 0.775\tiny{$\pm$0.010}  \\ 
			~ & Mean-pooling & 0.592\tiny{$\pm$0.026} & 0.597\tiny{$\pm$0.033} & 0.484\tiny{$\pm$0.029} & 0.685\tiny{$\pm$0.011} & 0.511\tiny{$\pm$0.022} & 0.797\tiny{$\pm$0.011}  \\ 
			~ & Clam-SB & 0.742\tiny{$\pm$0.024} & 0.763\tiny{$\pm$0.049} & {0.521\tiny{$\pm$0.046}} & 0.750\tiny{$\pm$0.039} & 0.514\tiny{$\pm$0.024} & 0.805\tiny{$\pm$0.017}
			\\ 
			~ & TransMIL & 0.643\tiny{$\pm$0.088} & 0.706\tiny{$\pm$0.076} & 0.444\tiny{$\pm$0.040} & 0.732\tiny{$\pm$0.043} & 0.385\tiny{$\pm$0.013} &	0.693\tiny{$\pm$0.027}
			\\
			&DSMIL & 0.736\tiny{$\pm$0.028} & 0.773\tiny{$\pm$0.034} & 0.511\tiny{$\pm$0.052} & 0.751\tiny{$\pm$0.028} & 0.458\tiny{$\pm$0.029} & 0.766\tiny{$\pm$0.023}
			\\
			&DTFD-MIL & 0.758\tiny{$\pm$0.051} & \underline{0.815\tiny{$\pm$0.063}} & 0.469\tiny{$\pm$0.016} & 0.717\tiny{$\pm$0.032} & 0.473\tiny{$\pm$0.021} & 0.776\tiny{$\pm$0.021}\\
			~ & IBMIL & \underline{0.777\tiny{$\pm$0.009}} &	0.799\tiny{$\pm$0.050} &	0.510\tiny{$\pm$0.043} &	0.726\tiny{$\pm$0.034} &	0.489\tiny{$\pm$0.017} &	0.791\tiny{$\pm$0.021}\\ 
			&MHIM-MIL  &0.752\tiny{$\pm$0.034}&	0.772\tiny{$\pm$0.026}&	0.511\tiny{$\pm$0.022}&	\textbf{0.774\tiny{$\pm$0.021}}&	\underline{0.543\tiny{$\pm$0.037}}&	\underline{0.816\tiny{$\pm$0.009}}\\
			&ABMIL&0.757\tiny{$\pm$0.020} &	0.790\tiny{$\pm$0.027} &	\underline{0.523}\tiny{$\pm$0.028} &	0.723\tiny{$\pm$0.035} &	0.465\tiny{$\pm$0.040} &	0.798\tiny{$\pm$0.013} \\
			\rowcolor{dino}&  ACMIL(ours)                    
			& \textbf{0.798\tiny{$\pm$0.029}} &	\textbf{0.841\tiny{$\pm$0.030}}        & \textbf{0.552\tiny{$\pm$0.048}} &	\underline{0.754\tiny{$\pm$0.008}} & \textbf{0.546\tiny{$\pm$0.028}} &	\textbf{0.821\tiny{$\pm$0.015}}   \\
			\midrule
			\multirow{9}{*}{\rotatebox{90}{\makecell{ViT-S/16 \\SSL pretrained}}} &
			Max-pooling & 0.903\tiny{$\pm$0.054} & 0.956\tiny{$\pm$0.029} & 0.596\tiny{$\pm$0.029} & 0.823\tiny{$\pm$0.033} & 0.590\tiny{$\pm$0.043} & 0.829\tiny{$\pm$0.023}  \\ 
			~ & Mean-pooling & 0.577\tiny{$\pm$0.057} & 0.569\tiny{$\pm$0.081} & 0.522\tiny{$\pm$0.038} & 0.739\tiny{$\pm$0.007} & 0.559\tiny{$\pm$0.024} & 0.827\tiny{$\pm$0.012}  \\ 
			~ & Clam-SB & 0.925\tiny{$\pm$0.035} & 0.969\tiny{$\pm$0.024} & 0.631\tiny{$\pm$0.034} & 0.863\tiny{$\pm$0.005} & 0.617\tiny{$\pm$0.022} & 0.865\tiny{$\pm$0.018}  \\ 
			~ & TransMIL & 0.922\tiny{$\pm$0.019} & 0.943\tiny{$\pm$0.009} & 0.631\tiny{$\pm$0.030} & 0.841\tiny{$\pm$0.006} & 0.539\tiny{$\pm$0.028} & 0.805\tiny{$\pm$0.010}  \\ 
			&DSMIL & 0.943\tiny{$\pm$0.007} & 0.966\tiny{$\pm$0.009} & 0.577\tiny{$\pm$0.028} & 0.816\tiny{$\pm$0.028} & 0.562\tiny{$\pm$0.028} & 0.820\tiny{$\pm$0.033} \\
			&DTFD-MIL & \underline{0.948\tiny{$\pm$0.007}} & \textbf{0.980\tiny{$\pm$0.011}} & 0.612\tiny{$\pm$0.080} & 0.870\tiny{$\pm$0.022} & 0.612\tiny{$\pm$0.034} & 0.842\tiny{$\pm$0.010}  \\
			~ & IBMIL & 0.912\tiny{$\pm$0.034} &	0.954\tiny{$\pm$0.022} &	{0.645\tiny{$\pm$0.041}} &	\underline{0.871\tiny{$\pm$0.014}} &	0.604\tiny{$\pm$0.032} &	0.834\tiny{$\pm$0.014} \\ 
			&MHIM-MIL                      
			&0.932\tiny{$\pm$0.024}&	0.970\tiny{$\pm$0.037}&	0.625\tiny{$\pm$0.060}&	0.865\tiny{$\pm$0.017}&	\underline{0.658\tiny{$\pm$0.041}}&	\underline{0.872\tiny{$\pm$0.022}}
			
			\\ 
			&ABMIL                     
			&0.914\tiny{$\pm$0.031} &	0.945\tiny{$\pm$0.027} &	\underline{0.680\tiny{$\pm$0.051}} &	0.866\tiny{$\pm$0.029} &	0.595\tiny{$\pm$0.036} &	0.831\tiny{$\pm$0.022} 
			\\
			\rowcolor{dino}& ACMIL(ours)                    
			& \textbf{0.954\tiny{$\pm$0.012}}             & \underline{0.974\tiny{$\pm$0.012}}             & \textbf{0.722\tiny{$\pm$0.030}} &	\textbf{0.888\tiny{$\pm$0.010}}
			& \textbf{0.662\tiny{$\pm$0.043}} &	\textbf{0.901\tiny{$\pm$0.011}}   \\
			\bottomrule
		\end{tabular}
		
		\label{tab2}
		\vspace{-4mm}
	\end{table*}
	\vspace{-2mm}
	\subsection{WSI Classification Results}\vspace{-1mm}

	Tab. \ref{tab2} provides a thorough comparison of performance between ACMIL and existing MIL methods. This evaluation spans three diverse datasets, involves two different choices for pretraining methods, and employs two crucial evaluation metrics, resulting in a comprehensive assessment with a total of 12 terms.
	
	Considering the overall performance, ACMIL consistently outshines existing methods. It secures the top position in 10 out of the 12 metrics and holds the second position in the remaining 2 metrics. Specifically, for the CAMELYON16, ACMIL achieves outstanding results using ResNet-18 pre-trained on ImageNet embeddings, surpassing the runner-up by 2.1\% and 2.6\% in terms of F1-score and AUC, respectively. On the other hand, with ViT-S/16 SSL pretrained embeddings, existing attention-based MIL methods exhibit remarkable performance, boasting F1-scores and AUC values exceeding 0.9. Notably, ACMIL achieves comparable performance with the former best-performing method, DTFD-MIL, in this setup.
	For the BRACS, ACMIL demonstrates a substantial lead when utilizing ViT-S/16 SSL pre-trained embeddings, surpassing the second-best performance by margins of 4.2\% and 1.7\% in F1-score and AUC, respectively. Moreover, when employing ResNet-18 pre-trained on ImageNet embeddings, ACMIL achieves comparable performance with the previously top-performing method, MHIM-MIL. For the LBC, ACMIL stands out significantly among the other methods across all four metrics.

	\begin{figure*}[t]
		\centering
		\includegraphics[width=0.95\textwidth]{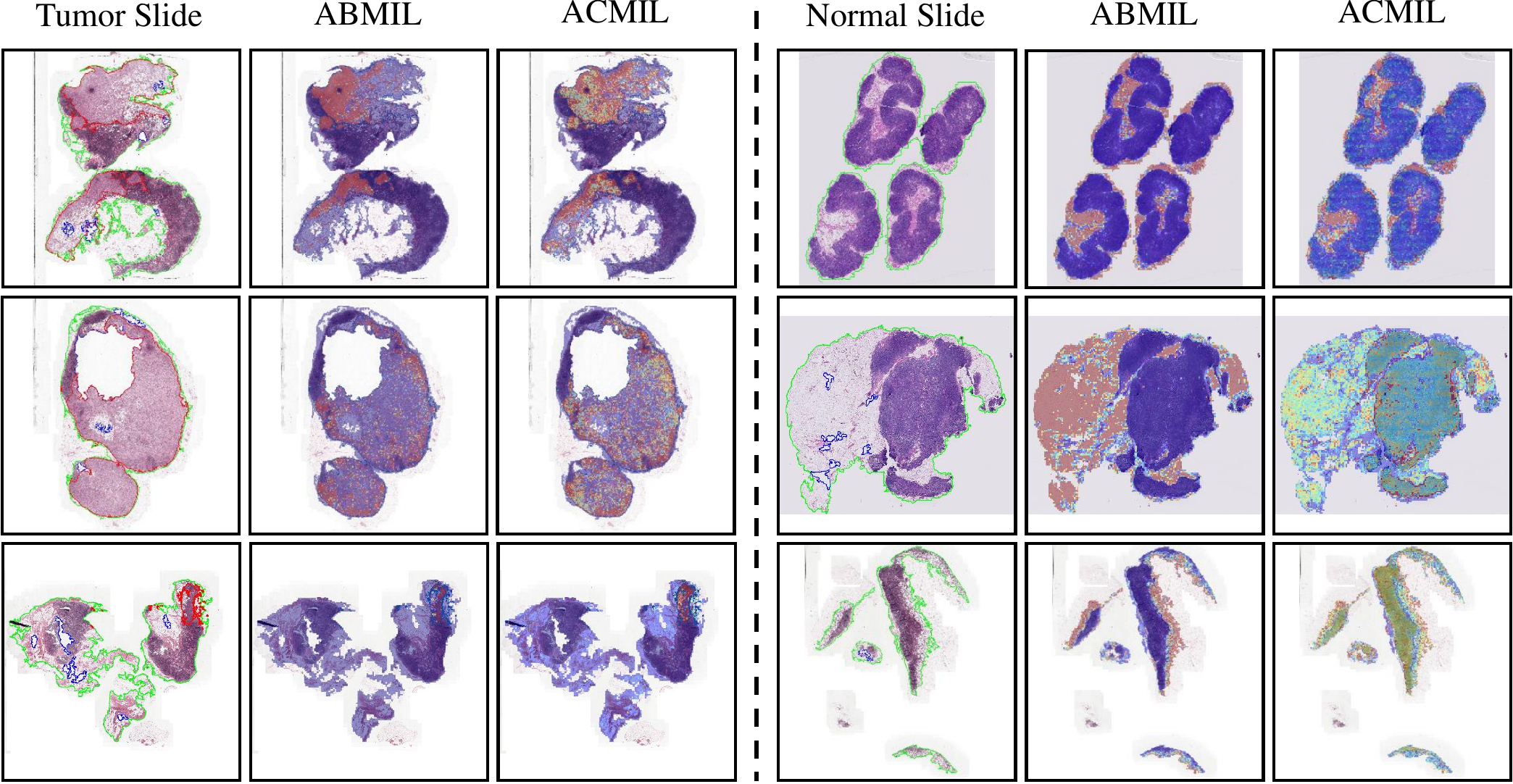}
		\vspace{-2mm}
		\caption{Heatmap visualization of WSI examples produced by ABMIL \cite{ilse2018attention} (baseline) and our ACMIL. The left part shows three tumor WSIs that come from CAMELYON16 dataset, and their tumor regions are delineated by red lines. ACMIL generates attention values that cover a more extensive portion of the tumor region compared to ABMIL. The right part shows three normal WSIs that come from CAMELYON16 dataset. ABMIL primarily focuses on a part of tissue such as adipose, while ACMIL extends its attention to the more normal tissues.}
		\label{fig_heatmap}
		\vspace{-4mm}
	\end{figure*}

	\subsection{Localization Results} \label{sec:localization}
	\textbf{Heatmap visualization.} Fig. \ref{fig_heatmap} presents heatmap visualizations illustrating examples of our approach's performance in comparison to the baseline method, ABMIL \cite{ilse2018attention}. Three tumor slides (left part) and three normal slides (right part) are selected to showcase the heatmap differences. Due to the space limitation, we present more visualization in Appendix Sec. \ref{sec_vis} for further insights. 
	
	For the tumor slides, ABMIL tends to concentrate its attention on only a fraction of the tumor regions, potentially overlooking other significant areas. In contrast, ACMIL allocates attention across a wider spectrum of tumor regions, resulting in better alignment with expert annotations.
	For the normal slides, ABMIL predominantly focuses on specific tissue types, such as adipose tissue. This will lead to misinterpretation that only the adipose tissue is the normal tissue and other normal regions are uncorrelated to the WSI label. On the other hand, ACMIL effectively distributes attention values to encompass all normal regions, ensuring all regions are correlated for the WSI label. This approach closely mimics human intuition and satisfies the definition of the MIL formulation.

	\begin{wraptable}{r}{0.4\textwidth}
		\vspace{-30pt}
		\caption{Comparison of FROC between ABMIL and ACMIL} 
		\tablestyle{8pt}{1.05}
		
		\begin{tabular}{x{30}|cc}
			&
			ABMIL
			&
			ACMIL
			\\
			\shline
			FROC & 0.3987 & 0.4233 \\
		\end{tabular}
		\label{tab:FROC}
		
		\vspace{-4mm}
	\end{wraptable}
	\noindent\textbf{FROC results.}  We employ the FROC metric suggested by CAMELYON16 challenge to evaluate the localization of tumor region quantitatively. As shown in Tab. \ref{tab:FROC}, the proposed ACMIL achieves higher FROC than ABMIL.

	\vspace{-2mm}
	\subsection{Ablation Study} \label{sec:ablation}
\begin{figure*}[t]
		\centering
		\includegraphics[width=0.95\textwidth]{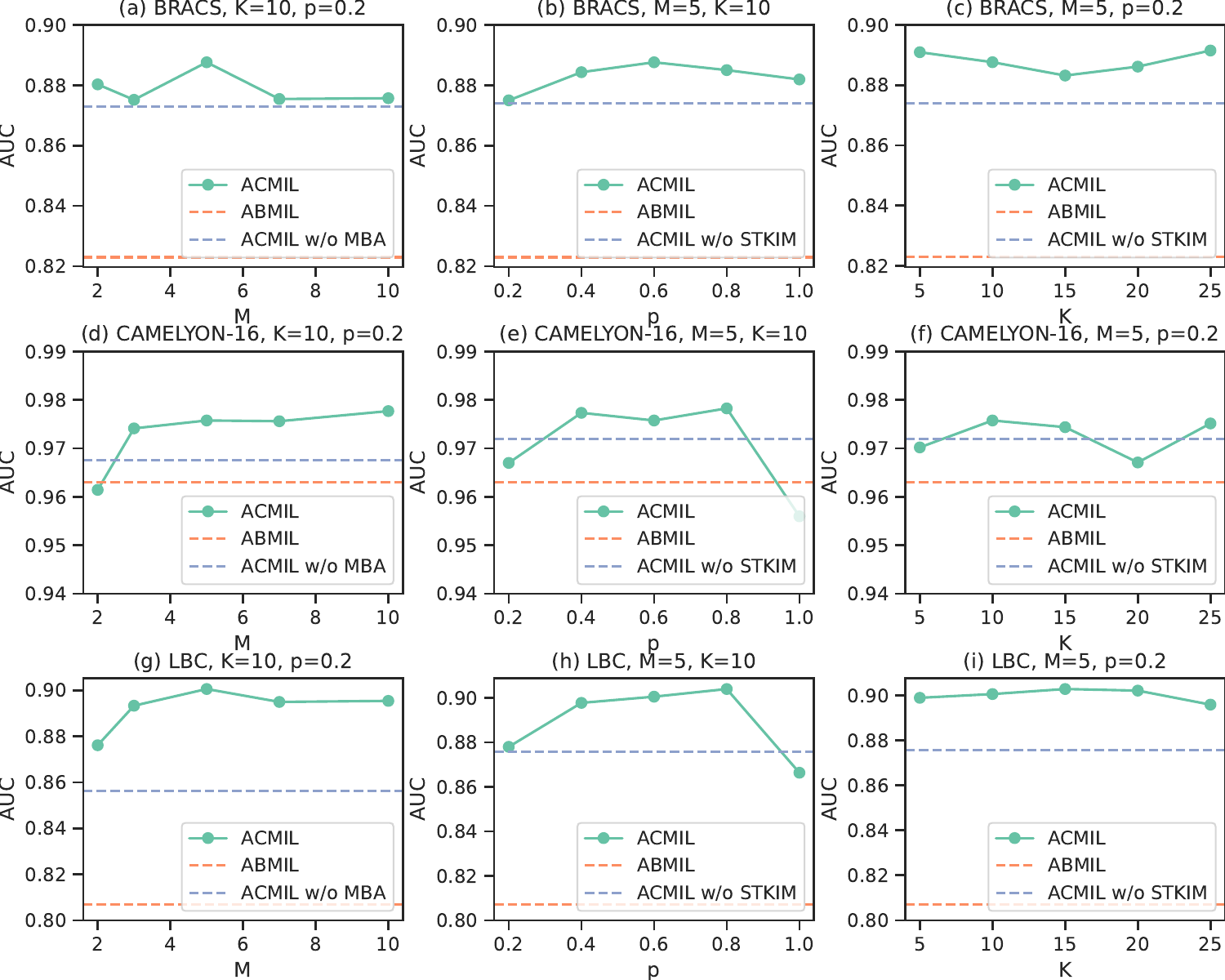}
		\caption{Ablation study on features extracted through the SSL pre-training. The effect of three hyperparameters, $K$, $p$, $M$, is investigated. Note that the orange dotted line denotes the performance of baseline, ABMIL, and the blue dotted line denotes the performance of ACMIL w/o MBA or STKIM. Five conclusions derived from the figure can be found in Sec. \ref{sec:ablation}.}
		\vspace{-2mm}
		\label{fig_ablation}
	\end{figure*}

\vspace{-1mm}
	Fig. \ref{fig_ablation} illustrates the AUC scores of ACMIL across three datasets when utilizing a ViT/B-16 feature extractor and varying hyperparameter settings. Several key observations emerge from these experiments:

	\noindent\textbf{Effect of branches number $M$ in MBA.} As shown in the first column, we find that the choice of $M$ affects performance significantly. Combining three datasets, setting $M=5$ achieves the best performance.  
	
	\noindent\textbf{Effect of masking probability $p$ in STKIM.} As shown in the second column, we find that the choice of $M$ also affects performance significantly. Notably, setting $p=1.0$ (masking all of Top-K instances) leads to performance deterioration across all three datasets. For LBC and CAMELYON, a $p=1.0$ setting even results in performance lower than the blue dotted lines.
	Otherwise, $p=0.6$ achieves the best performance on the BRACS dataset, whereas $p=0.8$ achieves the best performance on the other two datasets.
	
	\noindent\textbf{Effect of number of masking instances $K$ in STKIM.}  The third column shows that hyperparameter $K$ exhibits minimal sensitivity, where different $K$ values result in a performance difference of less than 1.0\% AUC. In practice, setting $K$ to 10 is generally sufficient for achieving near-optimal performance.
	
	\noindent\textbf{Implementing either MBA or STKIM individually leads to significant performance improvements.} The blue dotted lines represent ACMIL's AUC performance without MBA or STKIM, outperforming the orange dotted lines (ABMIL's performance) across all subfigures. Particularly noteworthy is the observation that MBA achieves better improvement than STKIM on all three datasets, with blue dotted lines in the last two columns surpassing those in the first column, especially on the CAMELYON and LBC datasets.
	
	\noindent\textbf{Combining MBA with STKIM yields greater performance improvements compared to using either MBA or STKIM alone.} The green dots represent ACMIL's performance under different hyperparameter combinations, with 39 out of 45 green dots exceeding blue dotted lines.
	
	\subsection{Further Analysis}\vspace{-1mm}
	
	\begin{figure}[t]
		\centering
		\begin{subfigure}[b]{0.46\textwidth}
			\includegraphics[width=\textwidth]{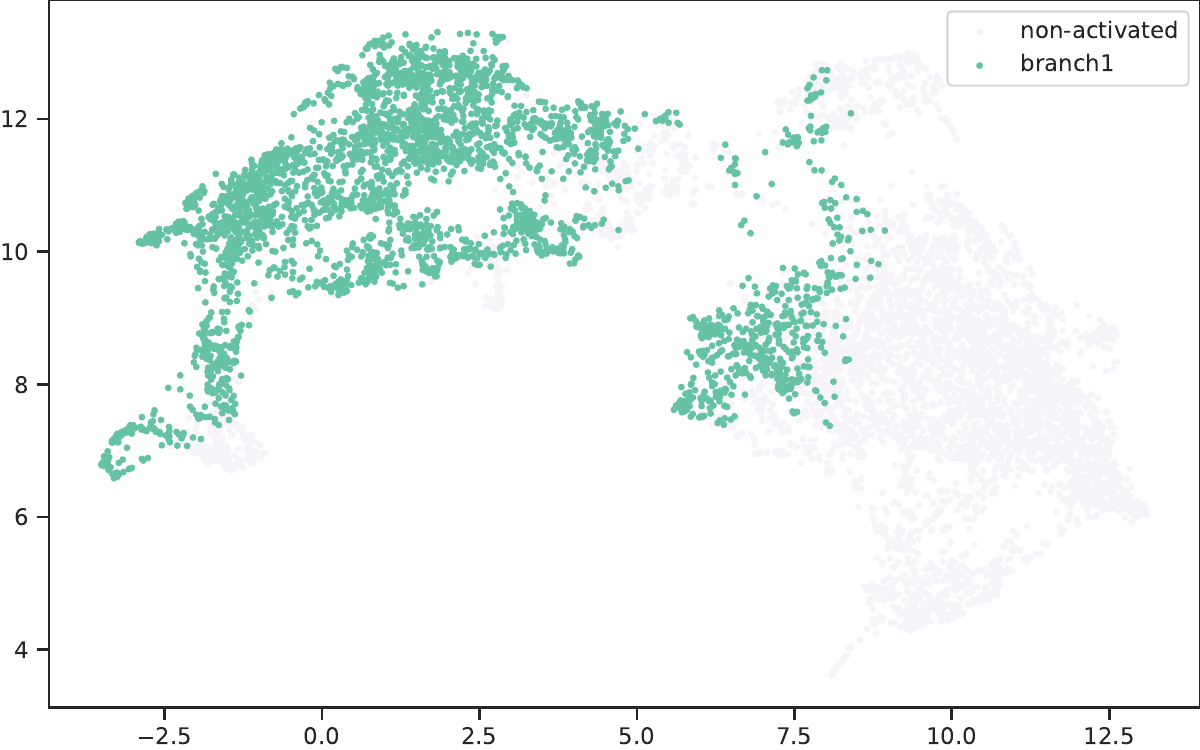}
			\caption{ABMIL}
			\label{figumap1}
		\end{subfigure}
		\begin{subfigure}[b]{0.46\textwidth}
			\includegraphics[width=\textwidth]{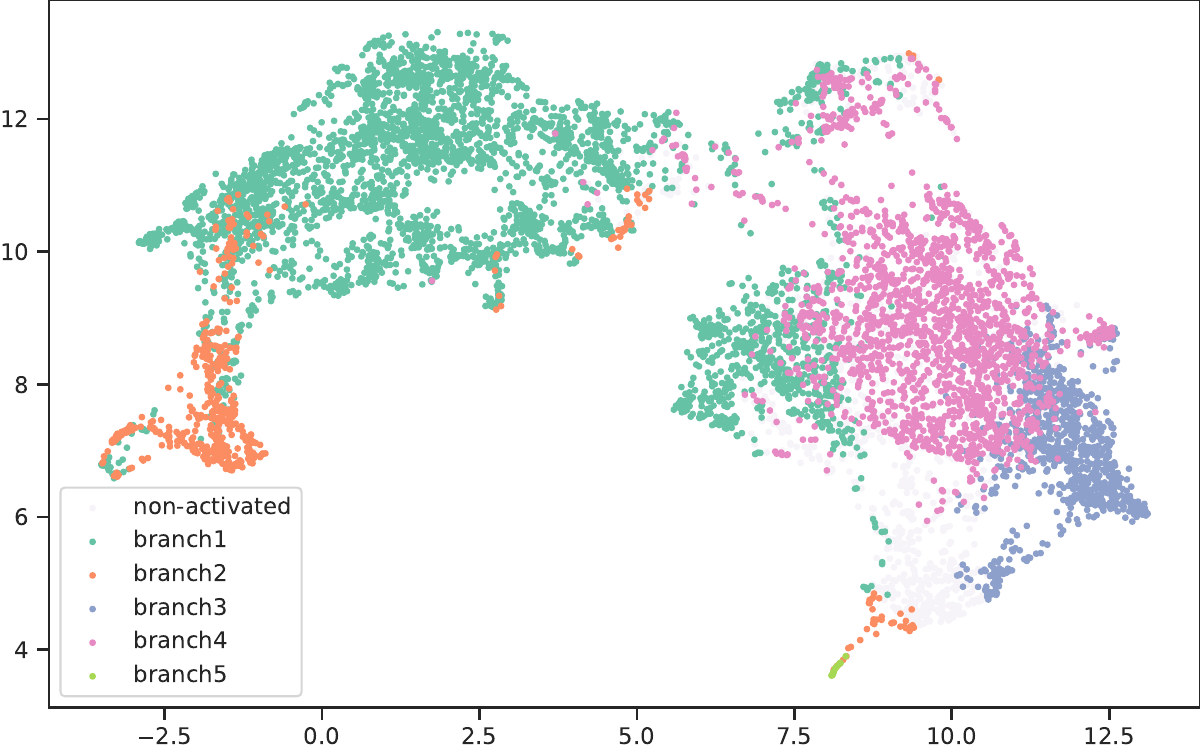}
			\caption{ABMIL with MBA}
			\label{figumap2}
		\end{subfigure}
		\vspace{-3mm}
		\caption{UMAP visualization of tumor instance features from the CAMELYON16 'test\_090' case.The tumor instances display distinct patterns, posing a challenge for a single branch to capture all of them. As a result, ABMIL overlooks the right pattern/cluster. In contrast, multiple branches in MBA capture different patterns separately, and their combination enables the activation of more tumor instances. An instance is considered active when its attention value surpasses $\frac{1}{N}$.}\label{figumap}
		\vspace{-3mm}
	\end{figure}

	\begin{figure}[t]
		\centering
		\begin{subfigure}[b]{0.46\columnwidth}
			\includegraphics[width=\columnwidth]{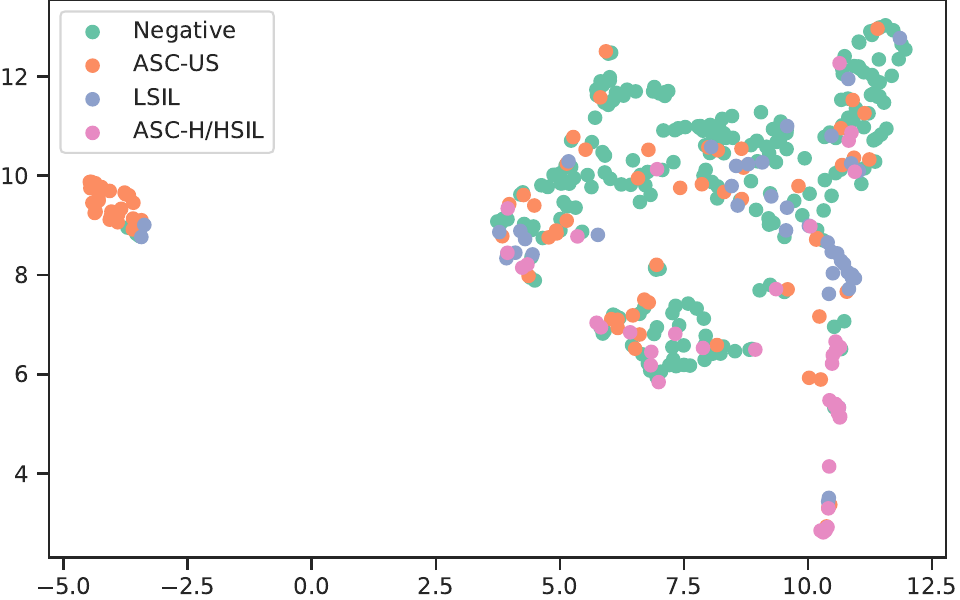}
			\caption{ABMIL(V-measure=0.224)}
			\label{figumapsw1}
		\end{subfigure}
		\begin{subfigure}[b]{0.46\columnwidth}
			\includegraphics[width=\columnwidth]{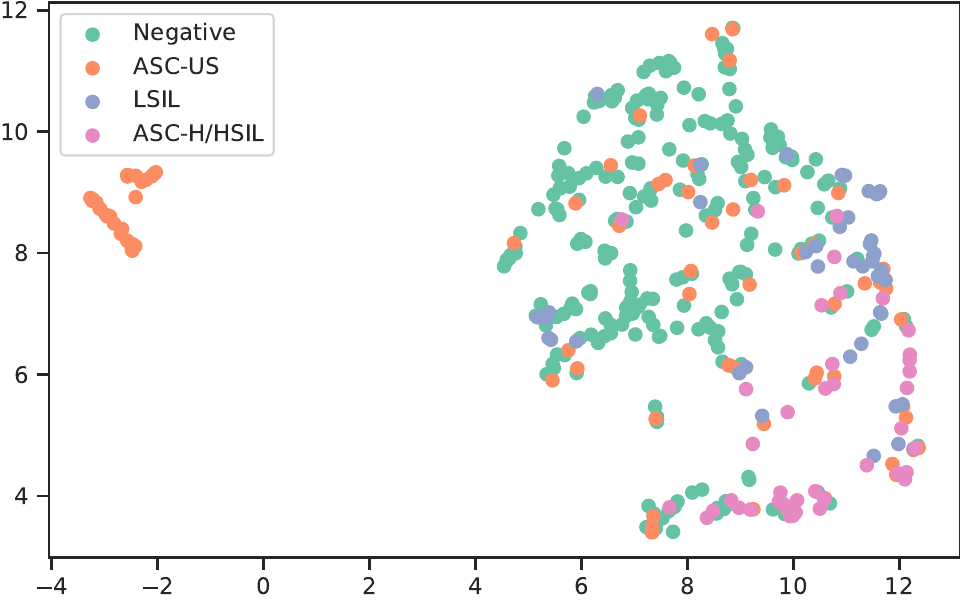}
			\caption{ACMIL(V-measure=0.316)}
			\label{figumapsw2}
		\end{subfigure}
		\vspace{-3mm}
		\caption{UMAP visualization \cite{mcinnes2018umap} of bag features for LBC test set. ACMIL effectively learns more discriminative features than ABMIL by improving the separation of  `LSIL' and `ASC-H/HSIL' features from the `Negative' class.  Improving feature separation is also corroborated by the V-measure score \cite{rosenberg2007v}, a clustering metric that considers both the homogeneity and completeness of the clusters.}\label{figumapsw}
		\vspace{-3mm}
	\end{figure}

	\textbf{MBA can capture diverse patterns.}  We employ UMAP \cite{mcinnes2018umap} to visualize instance features within the tumor region of the CAMELYON16 `test\_090' case. In Fig. \ref{figumap1}, it's evident that the tumor instances exhibit two primary patterns. However, ABMIL primarily activates the left pattern (colored orange) and neglects the right one. On the other hand, as demonstrated in Fig. \ref{figumap2}, MBA's various branches (branch1, branch2, branch3, and branch5) collectively capture the substructures of the left pattern, while branch4 specifically captures the right pattern. Combining all branches can capture more comprehensive patterns.

	\noindent\textbf{ACMIL can learn more discriminative bag features.} We employ UMAP \cite{mcinnes2018umap} to visualize bag features from the LBC test set, as illustrated in Fig. \ref{figumapsw1} and \ref{figumapsw2}. This visualization illustrates that our ACMIL is capable of learning more discriminative features compared to ABMIL. Specifically, ACMIL successfully separates the LSIL and ASC-H/HSIL clusters from the Negative cluster. To quantitatively assess the clustering performance, we employ V-measure \cite{rosenberg2007v}. ACMIL achieves a V-measure score of 0.316, a significant improvement over ABMIL, which scores 0.224.

	\begin{wrapfigure}{r}{0.5\textwidth}
		\centering
		\includegraphics[width=0.45\textwidth]{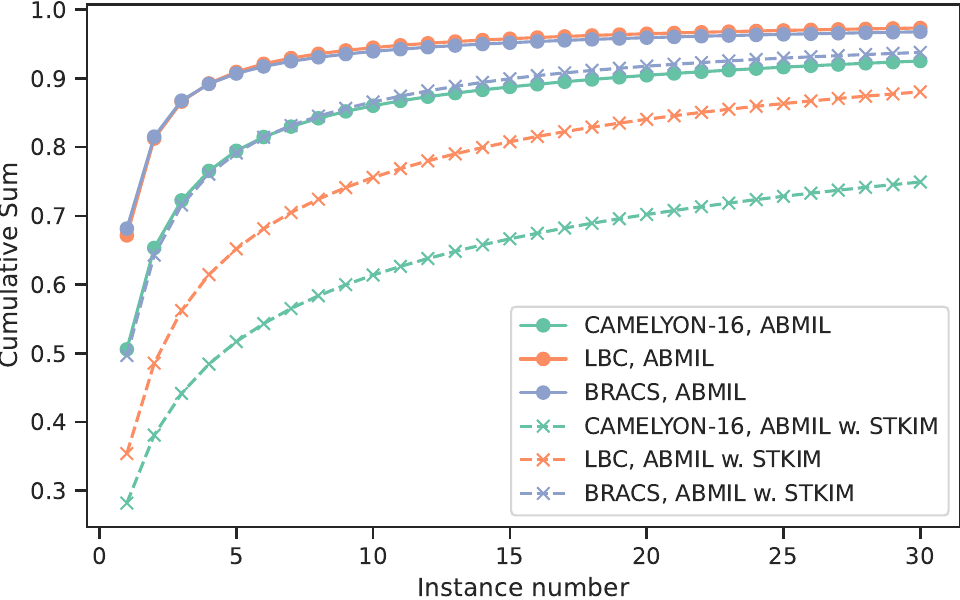}
		\vspace{-3mm}
		\caption{Comparison of the accumulative sum of Top-K attention values with and without STKIM. The use of STKIM helps alleviate the issue of excessive concentration of attention values within the Top-K range.}
		\label{fig_accum1}
		\vspace{-8mm}
	\end{wrapfigure}
	
	\noindent\textbf{STKIM can suppress the concentration of Top-K attention values.} Fig. \ref{fig_accum1} illustrates a comparison of the cumulative sum of Top-K attention values with and without STKIM. The plot clearly shows that the use of STKIM helps mitigate the scenario where Top-K attention values excessively dominate in the attention mechanism. This effect is particularly pronounced for CAMELYON16 dataset, where the cumulative sum of the top-10 values decreases from 0.87 to 0.6.

	\begin{table}[t] 
		\tiny
		\centering
		\caption{(a): Performance comparison between ACMIL with (w.) and without (w/o.) T-STKIM. T-STKIM means using STKIM at the test phase. The Gap column reports the performance difference between with and without T-STKIM. Using STKIM at the test phase slightly reduces performance. (b): Performance comparison between ACMIL with (w.) and without (w/o.) $\mathcal{L}_d$. The Gap column reports the performance difference between without and with $\mathcal{L}_d$. ACMIL without $\mathcal{L}_d$ drastically reduces its performance.}
		\begin{subtable}{0.47\linewidth} 
			\centering
			\vspace{-3mm}
			\caption{w. T-STKIM v.s. w/o. T-STKIM}
			\label{tab:T-sTKIM_ld_a}
			\vspace{-3mm}
			\begin{tabular}[t]{llccb}
				
				\toprule
				\multicolumn{5}{c}{ViT-S/16 SSL pre-trained} \\
				\midrule
				Dataset           & Metric & w. T-STKIM & w/o. T-STKIM & Gap(\%) \\ \midrule
				\multirow{2}{*}{Camelyon} & F1-score  & 0.927$\pm$0.057 & 0.954$\pm$0.012 & +2.7 \\
				& AUC  & 0.967$\pm$0.017 & 0.974$\pm$0.012 & +0.7 \\
				\multirow{2}{*}{BRACS} & F1-score & 0.697$\pm$0.033 & 0.722$\pm$0.030 & +2.5 \\
				& AUC & 0.875$\pm$0.012 & 0.888$\pm$0.010 & +1.3 \\
				\multirow{2}{*}{LBC} & F1-score & 0.637$\pm$0.034 & 0.662$\pm$0.043 & +2.5 \\
				& AUC & 0.878$\pm$0.012 & 0.901$\pm$0.011 & +2.3 \\
				\midrule
				\midrule
				\multicolumn{5}{c}{ResNet-18 Imagenet pre-trained } \\
				\midrule
				Dataset           & Metric & w. T-STKIM & w/o. T-STKIM & Gap(\%) \\ \midrule
				\multirow{2}{*}{Camelyon} & F1-score  & 0.780$\pm$0.026 & 0.798$\pm$0.029 & +1.8 \\
				& AUC  & 0.837$\pm$0.028 & 0.841$\pm$0.030 & +0.4 \\
				\multirow{2}{*}{BRACS} & F1-score & 0.566$\pm$0.054 & 0.552$\pm$0.048 & -1.4 \\
				& AUC & 0.750$\pm$0.021 & 0.754$\pm$0.008 & +0.4 \\
				\multirow{2}{*}{LBC} & F1-score & 0.535$\pm$0.027 & 0.546$\pm$0.028 & +1.1 \\
				& AUC & 0.808$\pm$0.019 & 0.821$\pm$0.015 & +1.3 \\
				
				\bottomrule
			\end{tabular}
			
		\end{subtable}
		\hfill
		\begin{subtable}{0.47\linewidth} 
			\centering
			\vspace{-3mm}
			\caption{w. $\mathcal{L}_d$ v.s. w/o. $\mathcal{L}_d$}
			\label{tab:T-sTKIM_ld_b}
			\vspace{-3mm}
			\begin{tabular}[t]{llccb}
				\toprule
				\multicolumn{5}{c}{ViT-S/16 SSL pre-trained} \\
				\midrule
				Dataset           & Metric & w/o. $\mathcal{L}_d$ & w. $\mathcal{L}_d$ & Gap(\%) \\ \midrule
				\multirow{2}{*}{Camelyon} & F1-score  & 0.901$\pm$0.037 & 0.954$\pm$0.012 & +5.3 \\
				& AUC  & 0.943$\pm$0.027 & 0.974$\pm$0.012 & +3.1 \\
				\multirow{2}{*}{BRACS} & F1-score & 0.642$\pm$0.046 & 0.722$\pm$0.030 & +8.0 \\
				& AUC & 0.859$\pm$0.020 & 0.888$\pm$0.010 & +2.9 \\
				\multirow{2}{*}{LBC} & F1-score & 0.603$\pm$0.023 & 0.662$\pm$0.043 & +5.9 \\
				& AUC & 0.837$\pm$0.009 & 0.901$\pm$0.011 & +6.4 \\
				\midrule
				\midrule
				\multicolumn{5}{c}{ResNet-18 Imagenet pre-trained } \\
				\midrule
				Dataset           & Metric & w/o. $\mathcal{L}_d$ & w. $\mathcal{L}_d$ & Gap(\%) \\ \midrule
				\multirow{2}{*}{Camelyon} & F1-score  & 0.747$\pm$0.022 & 0.798$\pm$0.029 & +5.1 \\
				& AUC  & 0.796$\pm$0.032 & 0.841$\pm$0.030 & +5.5 \\
				\multirow{2}{*}{BRACS} & F1-score & 0.500$\pm$0.031 & 0.552$\pm$0.048 & +5.2 \\
				& AUC & 0.760$\pm$0.026 & 0.754$\pm$0.008 & -0.6 \\
				\multirow{2}{*}{LBC} & F1-score & 0.532$\pm$0.019 & 0.546$\pm$0.028 & +1.4 \\
				& AUC & 0.809$\pm$0.018 & 0.821$\pm$0.015 & +1.2 \\
				
				\bottomrule
			\end{tabular}
		\end{subtable}
		\label{tab:T-sTKIM_ld}
	\end{table}

	\noindent\textbf{Do we need STKIM at the test phase? The answer is No.}
	In Tab. \ref{tab:T-sTKIM_ld_a}, we present the outcomes of ACMIL with and without STKIM during the test phase, along with the performance differences between these settings. Across 11 out of 12 evaluation metrics, the version of ACMIL without STKIM during testing outperforms the version with STKIM slightly. This suggests that STKIM is not necessary during the test phase. 
	
	\noindent\textbf{Do we need diversity loss in MBA? The answer is Yes.}
	In Tab. \ref{tab:T-sTKIM_ld_b}, we present the outcomes of ACMIL with and without $\mathcal{L}_d$, along with the performance differences between these settings. Notably, the last column clearly indicates a significant performance drop for ACMIL without $\mathcal{L}_d$. This emphasizes the crucial role of $\mathcal{L}_d$ in encouraging different branches to acquire distinctive discriminative knowledge within the MBA technique.

	\vspace{-1mm}
	\section{Conclusion}\vspace{-2mm}
	Due to the intrinsic properties of WSI, MIL methods have often led to overfitting, limiting their applications. This paper reveals that the overly concentrated attention values in the heatmap are closely related to overfitting. To address this, we propose ACMIL, which is underpinned by two novel techniques: MBA and STKIM. Our experimental results on three datasets demonstrate that ACMIL significantly surpasses SOTA methods. Moreover, this paper provides comprehensive experiments confirming the effectiveness of ACMIL in suppressing the attention value concentration and alleviating overfitting. We hope that our work can inspire future exploration into leveraging attention values for a comprehensive analysis of attention mechanisms. We also hope that our ACMIL can be applied to a broader spectrum of WSI analysis tasks.

	%
	%
	\bibliographystyle{splncs04}
	\bibliography{main}

	\section{Overview}
	In this appendix, we provide valuable resources and insights, including the source code (Sec. \ref{sec_sou}), implementation details (Sec. \ref{sec_imple}), additional experimental results (Sec. \ref{sec_res}), and a discussion on limitations (Sec. \ref{sec_limitation}). 
	
	\section{Source Code} \label{sec_sou}
	The source code of ACMIL is available at \url{https://github.com/dazhangyu123/ACMIL}. For further information on the environment setup and experiment execution, please refer to README.md. The implementation of ACMIL is based on the source code of ABMIL \cite{ilse2018attention} and CLAM \cite{lu2021data}. 
	
	\section{Implementation Details} \label{sec_imple}
	\noindent\textbf{Data Pre-processing.} We adopt the data pre-processing method from CLAM \cite{lu2021data}, which involves threshold segmentation and filtering to locate tissue regions in each whole-slide image (WSI). From these regions, we extract non-overlapping patches of size $256 \times 256$ at a magnification of $\times20$ for Camelyon16 and LBC datasets, and at a magnification of $\times10$ for BRACS.
	
	\noindent\textbf{Feature Extraction.} Given that ACMIL freezes the feature extractor during training, we extract and save features with 512 dimensions for ResNet-18 and 384 dimensions for ViT-S/16 to conserve space and expedite computation.
	
	\noindent\textbf{Model Architecture.} The learnable components of the model include one fully-connected layer to reduce features to 256 dimensions for ResNet-18 and 128 dimensions for ViT-S/16, a gated attention network, and a fully-connected layer for making predictions.
	
	\noindent\textbf{Training.} All models are trained for 100 epochs using a cosine learning rate decay starting at 0.0001 for ViT-S/16 and 0.0002 for ResNet-18. We employ an Adam optimizer with a weight decay of 0.0001, and the batch size is set to 1.
	
	\noindent\textbf{Hyperparameters.} For the setting of Camelyon16 and natural supervised pre-training, we set hyperparameters as $M=2, K=10, p=0.6$. For the other situation,  we set hyperparameters as  $M=5, K=10, p=0.6$. 
	
	\section{More Experimental Results} \label{sec_res}
	The additional experimental results include performance comparison against baselines (Sec. \ref{sec_base}), visualization of validation metrics (Sec. \ref{sec_metric}),  additional heatmap visualizations (Sec. \ref{sec_vis}), UMAP visualization of normal instances (Sec. \ref{sec_feat}), performance comparison between different masking strategies (Sec. \ref{sec_mask}), and discussion about computational cost  (Sec. \ref{sec_dis}). 
	
	\begin{table*}[t]
		\scriptsize
		\centering
		\caption{{The performance comparison between the baseline and our ACMIL across two attention mechanisms (i.e., gated attention (GA) and multiple head attention (MHA)), three datasets, and two pretrained methods.}}
		\setlength{\tabcolsep}{0.5mm}{\begin{tabular}{lccccccc}
				\toprule
				
				\multirow{2}{*}{\diagbox[width=8em]{{Method}}{{Performance}}} & \multicolumn{2}{c}{CAMELYON-16} & \multicolumn{2}{c}{BRACS} &  \multicolumn{2}{c}{LBC} &  \multirow{2}{*}{Average}\\ 
				\cmidrule(lr){2-3} \cmidrule(lr){4-5} \cmidrule(lr){6-7}
				& F1-score       & AUC           & F1-score       & AUC       & F1-score            & AUC  & \\\midrule
				\multicolumn{8}{c}{ResNet18 ImageNet pretrained} \\  \midrule           
				
				GA                     
				&0.757\tiny{$\pm$0.020} &	0.790\tiny{$\pm$0.027} &	0.523\tiny{$\pm$0.028} &	0.723\tiny{$\pm$0.035} &	0.465\tiny{$\pm$0.040} &	0.798\tiny{$\pm$0.013} &  0.676\\
				+ACMIL                     
				& {0.798\tiny{$\pm$0.029}} &	{0.841\tiny{$\pm$0.030}}        & {0.552\tiny{$\pm$0.048}} &	{0.754\tiny{$\pm$0.008}} & {0.546\tiny{$\pm$0.028}} &	{0.821\tiny{$\pm$0.015}} & 0.719   \\
				${\Delta}(\%)$
				& \color{red}{+4.1}
				& \color{red}{+5.1}
				& \color{red}{+2.9}
				& \color{red}{+3.1}
				& \color{red}{+8.1}
				& \color{red}{+2.3} 
				& \color{red}{+4.3} \\
				MHA                  
				& 0.752\tiny{$\pm$0.030} &	0.775\tiny{$\pm$0.027} &	0.502\tiny{$\pm$0.039} &	0.738\tiny{$\pm$0.019} &	0.531\tiny{$\pm$0.025} &	0.817\tiny{$\pm$0.011} & 0.686 \\
				
				+ACMIL                     
				& 0.799\tiny{$\pm$0.018} &	0.875\tiny{$\pm$0.017} &	0.541\tiny{$\pm$0.063} &	0.723\tiny{$\pm$0.028} &	0.555\tiny{$\pm$0.038} &	0.818\tiny{$\pm$0.012} & 0.719
				
				\\
				$\Delta(\%)$
				& \color{red}{+4.7}
				& \color{red}{+10.0}
				& \color{red}{+3.9}
				& \color{gray}{-1.5}
				& \color{red}{+2.4}
				& \color{red}{+0.1}
				& \color{red}{+3.3}\\
				\midrule
				\multicolumn{8}{c}{ViT-S/16 SSL pretrained} \\   \midrule          
				
				GA                     
				&0.914\tiny{$\pm$0.031} &	0.945\tiny{$\pm$0.027} &	0.680\tiny{$\pm$0.051} &	0.866\tiny{$\pm$0.029} &	0.595\tiny{$\pm$0.036} &	0.831\tiny{$\pm$0.022} & 0.805
				\\
				+ACMIL                     
				& {0.954\tiny{$\pm$0.012}}             & {0.974\tiny{$\pm$0.012}}             & {0.722\tiny{$\pm$0.030}} &	{0.888\tiny{$\pm$0.010}}& {0.662\tiny{$\pm$0.043}} &	{0.901\tiny{$\pm$0.011}} & 0.850  \\
				${\Delta}(\%)$
				& \color{red}{+4.0}
				& \color{red}{+2.9}
				& \color{red}{+4.2}
				& \color{red}{+2.2}
				& \color{red}{+6.7}
				& \color{red}{+7.0}
				& \color{red}{+4.5}\\
				MHA                  
				& 0.931\tiny{$\pm$0.032} &	0.961\tiny{$\pm$0.017} &	0.656\tiny{$\pm$0.030} &	0.850\tiny{$\pm$0.030} &	0.619\tiny{$\pm$0.032} &	0.864\tiny{$\pm$0.013} & 0.813\\
				
				+ACMIL                     
				&	0.936\tiny{$\pm$0.027} &	0.973\tiny{$\pm$0.014} &	0.667\tiny{$\pm$0.059} &	0.879\tiny{$\pm$0.028} &	0.649\tiny{$\pm$0.024} &	0.876\tiny{$\pm$0.012} & 0.830
				\\
				
				$\Delta(\%)$
				& \color{red}{+0.5}
				& \color{red}{+1.2}
				& \color{red}{+1.1}
				& \color{red}{+2.9}
				& \color{red}{+3.0}
				& \color{red}{+1.2}
				& \color{red}{+1.7}\\
				
				\bottomrule
		\end{tabular}}
		\label{tab1}
	\end{table*}

	\subsection{Performance Evaluation against Baselines} \label{sec_base}
	
	To assess the adaptability of our ACMIL to different attention mechanisms, we selected two prominent attention mechanisms as our baselines. The first is the gated attention (GA) mechanism \cite{dauphin2017language}, employed in approaches like ABMIL \cite{ilse2018attention}, CLAM \cite{lu2021data}, and DTFD-MIL \cite{zhang2022dtfd}. The second is the multiple head attention (MHA) mechanism \cite{vaswani2017attention}, utilized in methods such as TransMIL \cite{shao2021transmil} and IPS transformer \cite{bergner2022iterative}. The results are presented in Table \ref{tab1}.
	
	With GA as the baseline, ACMIL exhibits a substantial and comprehensive improvement in performance. All 12 performance metrics show enhancements, with an average gain of 4.4 points, a minimum increase of 2.2 points, and a maximum improvement of 8.1 points.
	
	With MHA as the baseline, ACMIL also demonstrates performance improvements in the majority of terms (i.e., 11 out of 12 terms), achieving an average improvement of 2.5 points. In comparison to GA, MHA introduces parallel processing (i.e., heads). This modification enables the learning of different visual concepts across heads \cite{chen2022scaling}, contributing to a slight attenuation in the improvements brought by ACMIL.

	\subsection{Visualization of validation metrics across training epochs} \label{sec_metric} \label{vis_metric}
	\begin{figure}[t]
		\centering
		\begin{subfigure}[b]{0.48\columnwidth}
			\includegraphics[width=\textwidth]{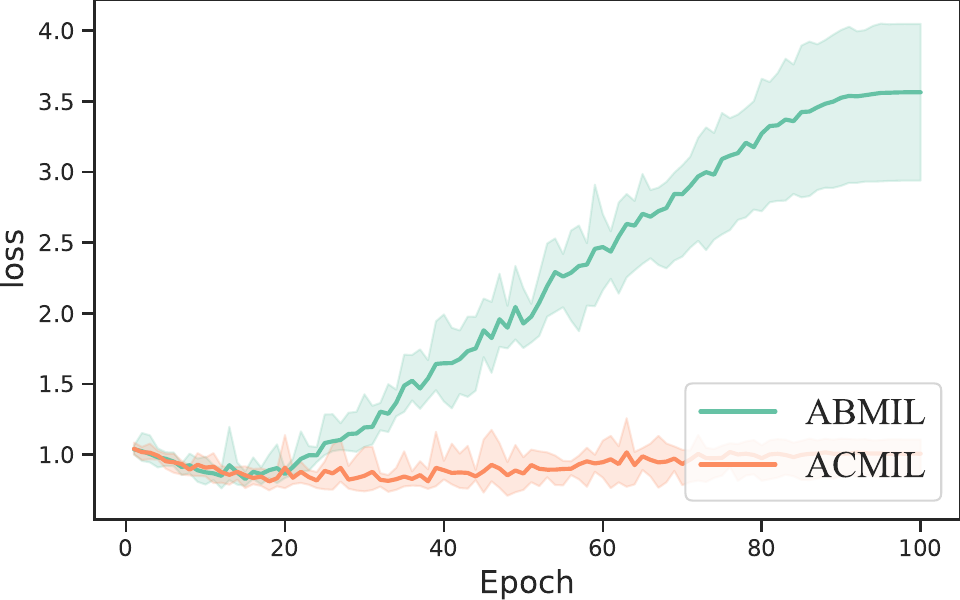}
			\caption{Validation Loss}
			\label{fig11}
		\end{subfigure}
		\begin{subfigure}[b]{0.48\columnwidth}
			\includegraphics[width=\textwidth]{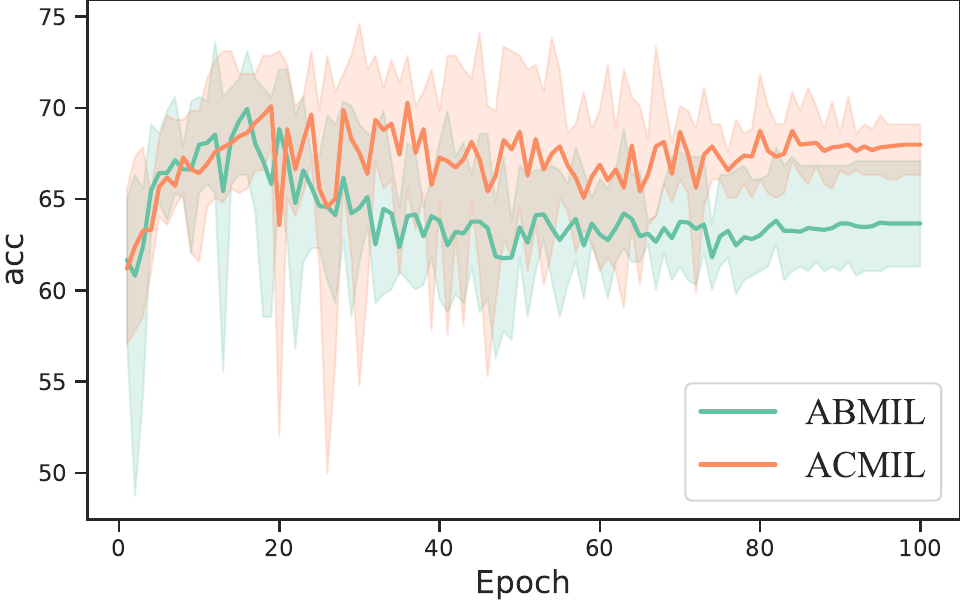}
			\caption{Validation Acc.}
			\label{fig12}
		\end{subfigure}
		\begin{subfigure}[b]{0.48\columnwidth}
			\includegraphics[width=\textwidth]{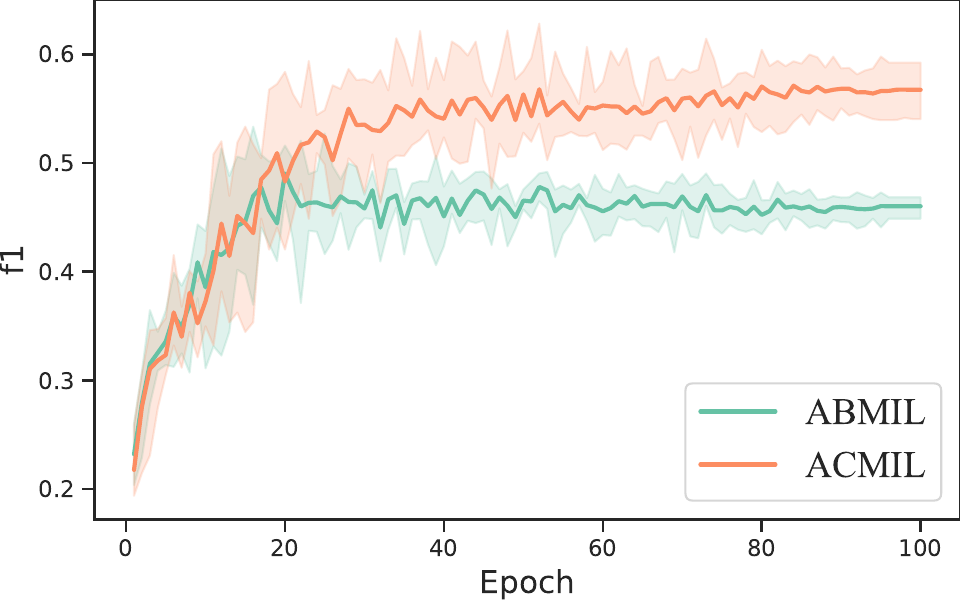}
			\caption{Validation F1-score}
			\label{fig13}
		\end{subfigure}
		\begin{subfigure}[b]{0.48\columnwidth}
			\includegraphics[width=\textwidth]{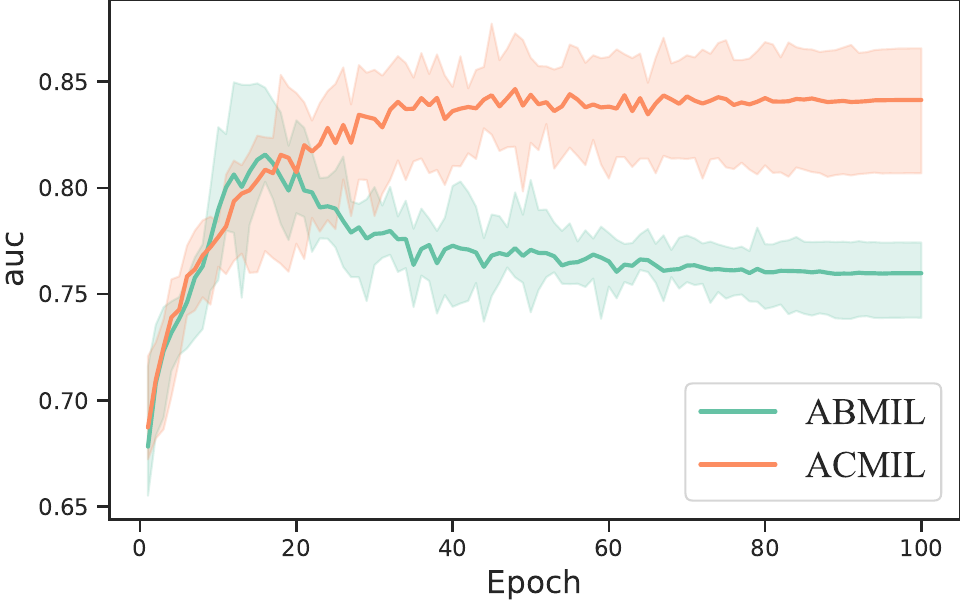}
			\caption{Validation AUROC}
			\label{fig14}
		\end{subfigure}
		\caption{Performance comparison between ABMIL \cite{ilse2018attention} and our ACMIL on LBC validation set throughout the training process. ABMIL displays pronounced signs of overfitting, as indicated by a significant increase in validation loss and a decline in the other three evaluation metrics. Conversely, ACMIL effectively mitigates the overfitting issue.}\label{fig:valid}
	\end{figure}
	
	Fig. \ref{fig:valid} depicts validation metrics across training epochs. ABMIL \cite{ilse2018attention}, one of the most commonly-used MIL methods, shows significant overfitting since loss drastically increases and validation metrics significantly decrease as the training processes. As a comparison, ACMIL suppresses the increase of validation loss throughout the training process and the decrease of the other three evaluation metrics. As a result, ACMIL indeed effectively mitigates the overfitting issue.

	\begin{figure*}[t]
		\centering
		\includegraphics[width=0.85\columnwidth]{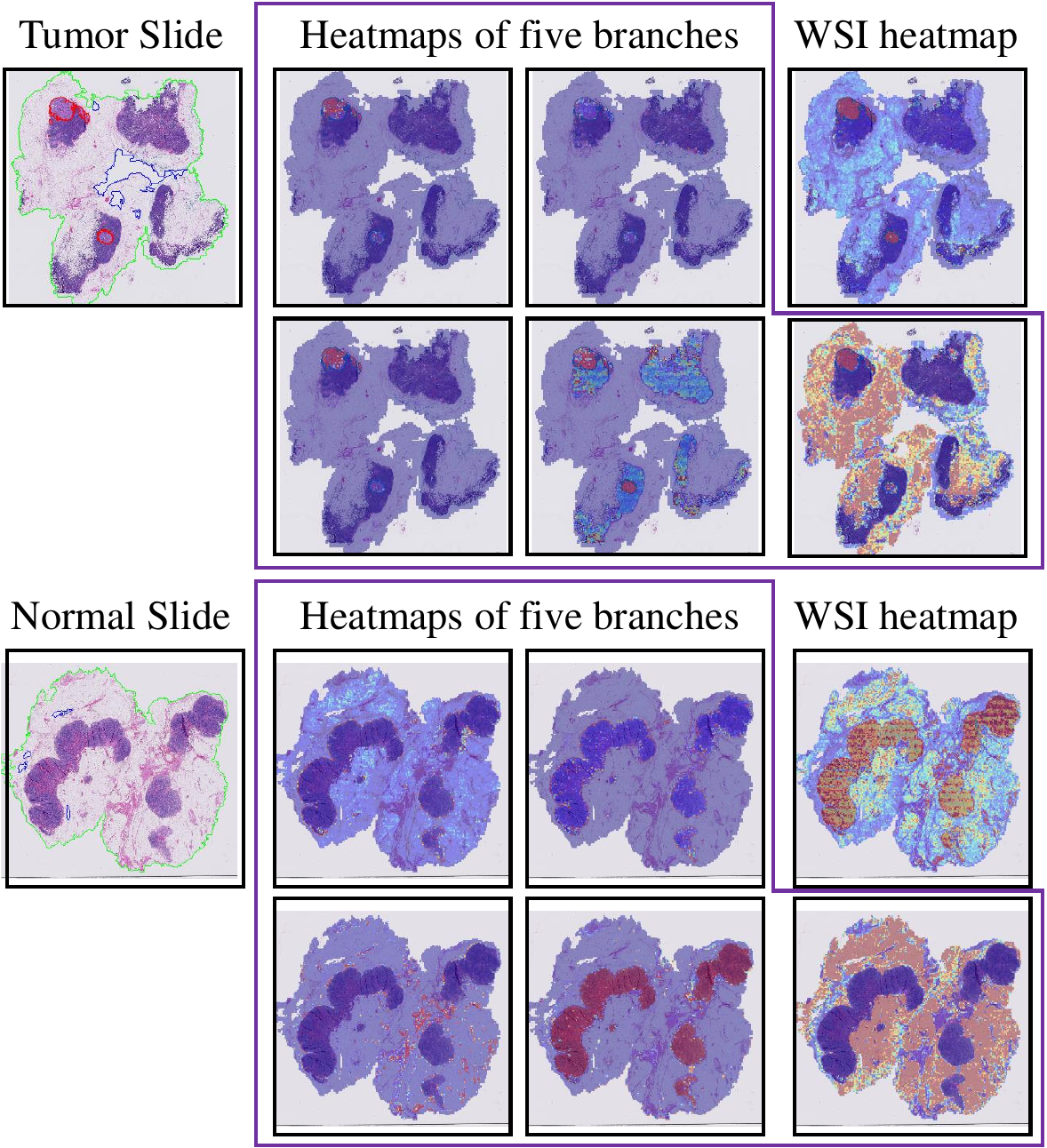}
		\caption{Heatmap visualizations for five attention branches. Different branches specialize in capturing specific features, contributing to better interpretability for the bag (final) heatmap.}
		\label{fig_branch}
	\end{figure*}
	
	\subsection{More heatmap visualization} \label{sec_vis}

	\noindent\textbf{Heatmap visualizations of five attention branches in MBA.} In Fig. \ref{fig_branch}, we present the heatmap visualizations for five attention branches and delve into the effects of these distinct branches. We've chosen two test slides in Camelyon16 for this analysis, including one tumor slide and one normal slide. For the tumor slide, we observe that all five branches capture the cancerous instances. Notably, the third and fifth branches successfully capture the entirety of the tumor regions, while the remaining three branches only manage to capture a subset of the tumor regions. Additionally, the third branch activates the adipose, and the fifth branch activates the lymphocyte regions. Overall, the averaged heatmap captures the whole tumor regions, along with slightly activating some normal regions. For the normal slide, the first two branches activate instances lying between adipose and lymphocyte regions. The third branch predominantly activates adipose tissue, the fourth branch emphasizes muscle regions, and the fifth branch highlights lymphocyte regions. Overall, the averaged heatmap activates all normal regions. This analysis illustrates how the different branches specialize in capturing specific features, contributing to a more comprehensive understanding of the data.
	
	\begin{figure}[t]
		\centering
		\includegraphics[width=0.9\columnwidth]{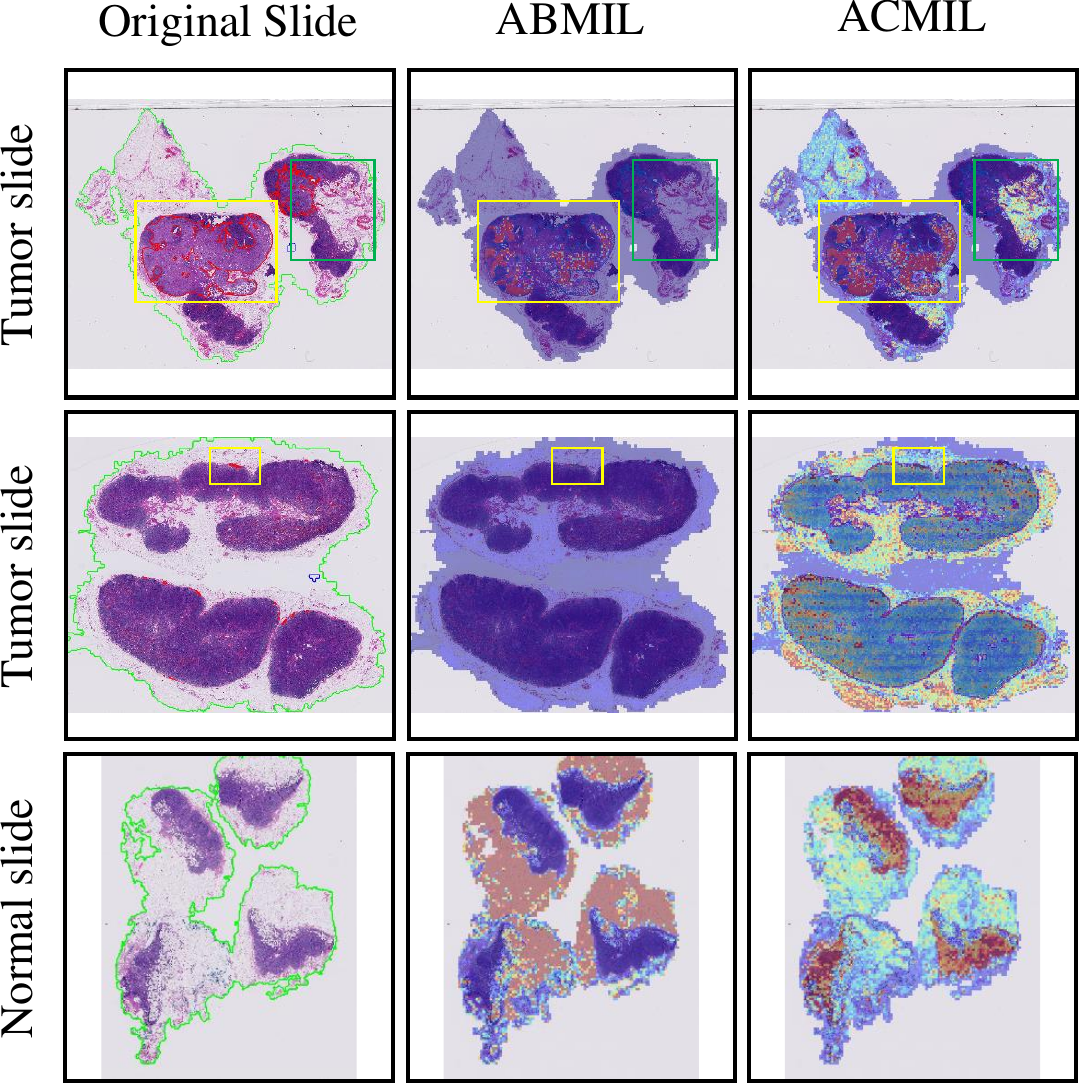}
		\caption{Heatmap visualizations with bad interpretability. Three cases indicate that ACMIL's approach of assigning broader attention values to a wide range of predictive instances doesn't consistently enhance interpretability.}
		\label{fig_fail}
	\end{figure}

	\noindent\textbf{Heatmap visualizations with bad interpretability.} In Fig. \ref{fig_fail}, we present three cases (i.e., two tumor slides and one normal slide) with heatmap visualizations that exhibit poor interpretability.  The first slide is a tumor slide. While ACMIL activates a greater number of cancerous instances than ABMIL (as indicated by the yellow box), it also activates some normal instances (visible in the green box). This mixed activation can potentially mislead experts during practical interpretability analysis. The second instance also concerns a tumor case but with small tumor regions. ABMIL accurately localizes the tumor regions (see yellow box). In contrast, ACMIL allocates more attention values to a broader range of predictive instances, which results in an inability to precisely locate the tumor regions. The third case pertains to a normal slide. In contrast to ABMIL, which provides misleading interpretability by predominantly focusing on adipose tissue, ACMIL assigns excessive attention values to lymphocyte regions. Consequently, the heatmap primarily highlights lymphocyte tissue instead of the expected comprehensive representation of normal tissue.
	
	\begin{figure}[t]
		\centering
		\includegraphics[width=\columnwidth]{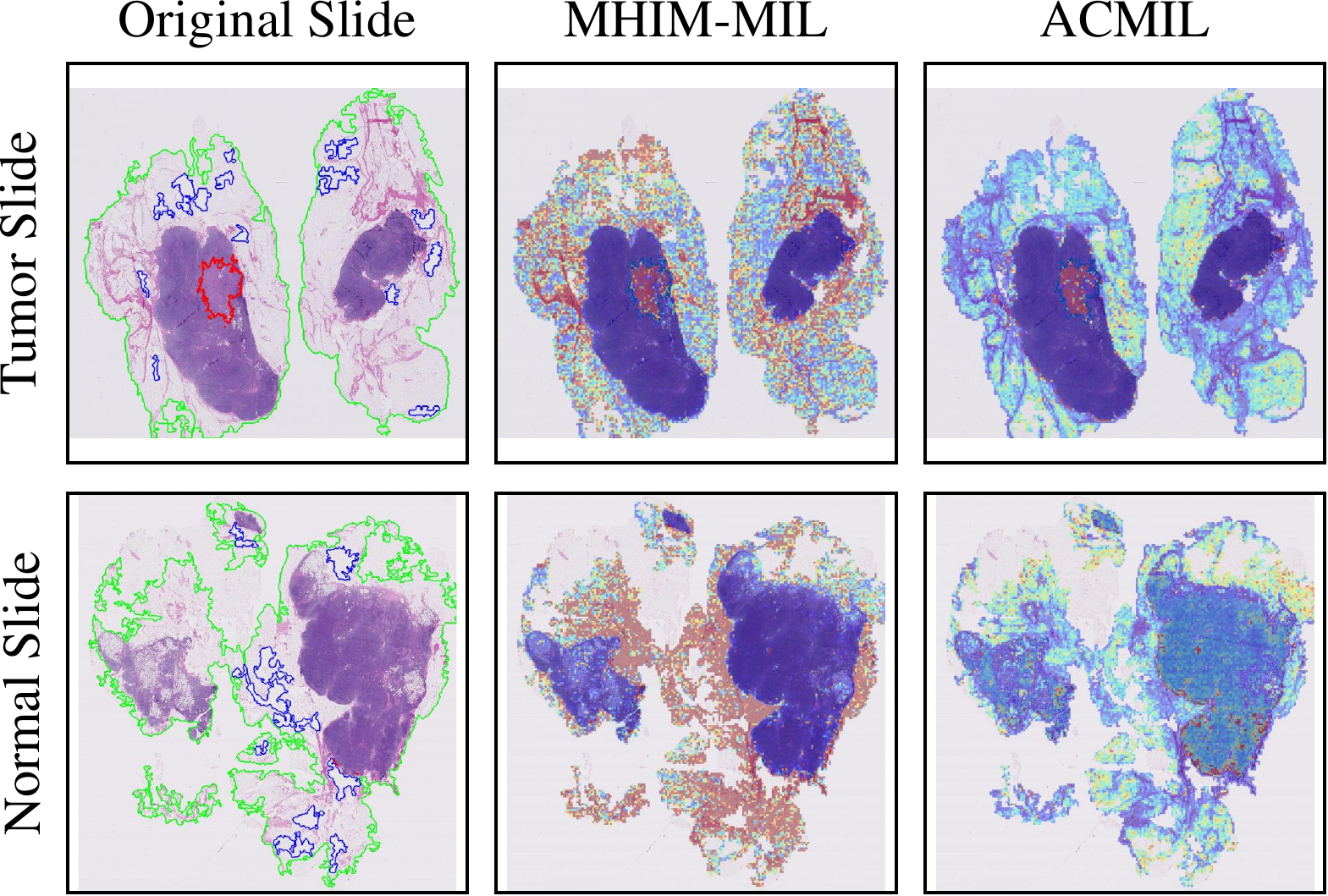}
		\caption{Comparison of heatmap visualizations between  MHIM-MIL \cite{tang2023multiple} and ACMIL (Zoom in for best view). ACMIL performs better in capturing comprehensive predictive instances.}
		\label{fig_mhim}
	\end{figure}
	
	\noindent\textbf{Comparison of heatmap visualization between MHIM-MIL \cite{tang2023multiple} and ACMIL.} In Fig. \ref{fig_mhim}, we present the heatmap visualizations of MHIM-MIL and ACMIL. For the tumor slide (first row), MHIM-MIL and ACMIL both capture all cancerous instances in the tumor region, but MHIM-MIL activates more normal instances than ACMIL. For the normal slide (second row), MHIM-MIL predominately activates adipose, whereas ACMIL activates all normal instances more uniformly.
	
	\begin{figure*}[t]
		\centering
		\begin{subfigure}[b]{0.48\textwidth}
			\includegraphics[width=\textwidth]{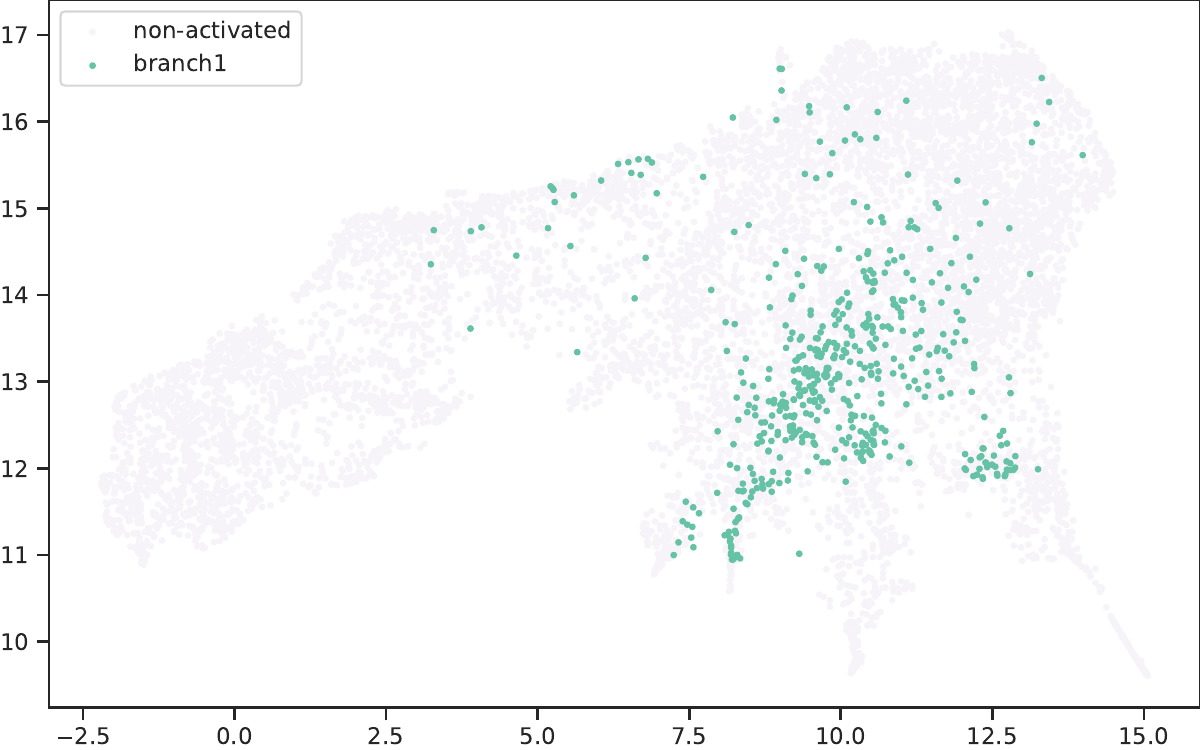}
			\caption{ACMIL without MBA}
			\label{figumapu1}
		\end{subfigure}
		\begin{subfigure}[b]{0.48\textwidth}
			\includegraphics[width=\textwidth]{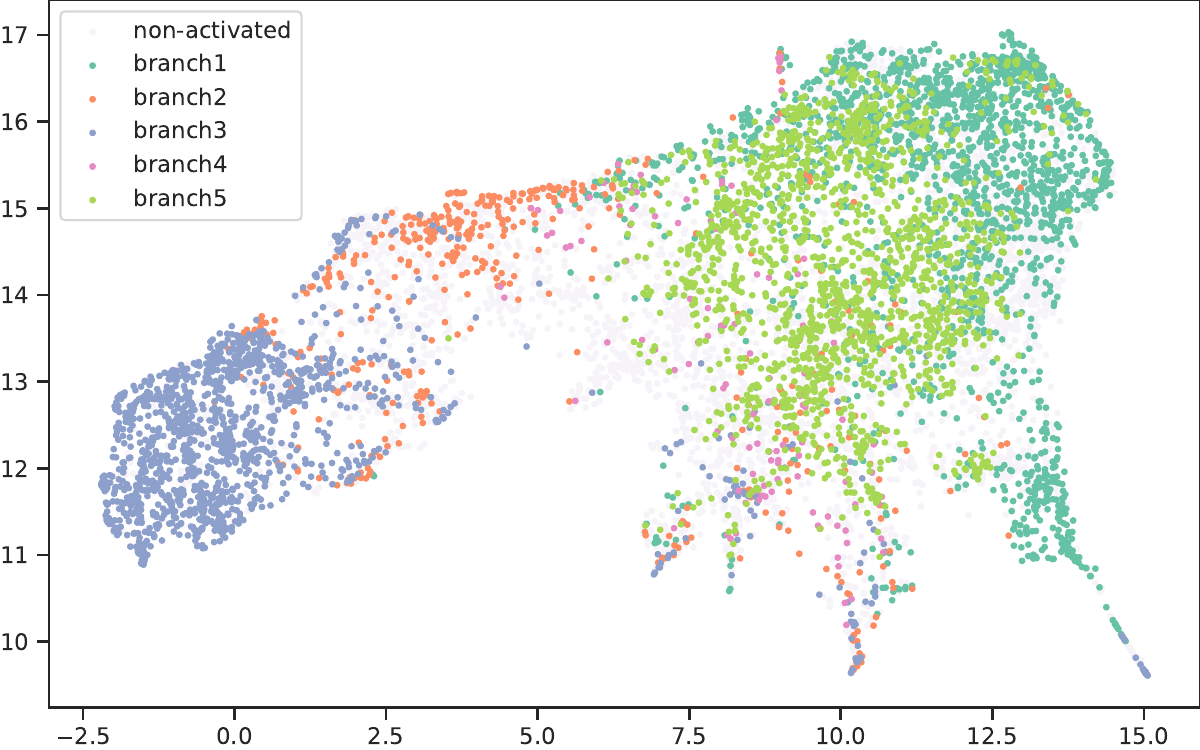}
			\caption{ACMIL}
			\label{figumapu2}
		\end{subfigure}
		\caption{UMAP visualization \cite{mcinnes2018umap} of instance features in a normal case, Camelyon16 'test\_016'. The normal instances exhibit distinct patterns, making it challenging for a single-branch model like ABMIL to capture them comprehensively. Consequently, ABMIL may overlook certain instances. In contrast, our ACMIL leverages multiple branches, each adept at capturing specific patterns, enabling ACMIL to activate a greater number of normal instances. Note that the instance is considered active when its attention value surpasses $\frac{1}{N}$.}\label{figumapn}
	\end{figure*}
	
	\subsection{Instance Feature Analysis for Normal Slide}  \label{sec_feat}
	
	In Fig. \ref{figumapn}, we present the UMAP visualization \cite{mcinnes2018umap} of normal instance features in a typical Camelyon case, 'test\_016'. The comparison between ABMIL and ACMIL is quite evident. ABMIL, with a single attention branch, activates only a fraction of normal instances. Conversely, ACMIL utilizes five branches, with each branch specializing in capturing specific patterns, resulting in the activation of nearly all normal instances. This observation demonstrates the superior ability of ACMIL to encompass a broader range of patterns in the data.
	
	\begin{table*}[t]
		\scriptsize
		\centering
		\caption{{The performance comparison between different masking strategies on SSL pre-trained embedding. Strategy1 denotes the default masking strategy in STKIM, i.e., masking 10 instances with a probability of 0.6. Strategy2 denotes the default masking strategy in WENO \cite{qu2022bi}, i.e., masking 95 instances with a probability of 1.0.  Strategy3 denotes the default masking strategy in MHIM-MIL \cite{tang2023multiple}, i.e., masking 1\% instances with a probability of 0.5. Strategy 1 outperforms the other two strategies.}}
		\setlength{\tabcolsep}{1.0mm}{\begin{tabular}{lcccccc}
				\toprule
				
				\multirow{2}{*}{\diagbox[width=8em]{{Method}}{{Performance}}} & \multicolumn{2}{c}{CAMELYON-16} & \multicolumn{2}{c}{BRACS} &  \multicolumn{2}{c}{LBC} \\ 
				\cmidrule(lr){2-3} \cmidrule(lr){4-5} \cmidrule(lr){6-7}
				& F1-score       & AUC           & F1-score       & AUC       & F1-score            & AUC   \\\midrule       
				
				Strategy1                    
				& 0.954          & {0.974}             & {0.722} &	{0.888}& {0.662} &	{0.901}  \\
				Strategy2                  
				& {0.741\color{red}{\tiny{(-0.213)}}} &	{0.843\color{red}{\tiny{(-0.131)}}}        & {0.521\color{red}{\tiny{(-0.201)}}} &	{0.844\color{red}{\tiny{(-0.044)}}} & {0.502\color{red}{\tiny{(-0.160)}}} &	{0.784\color{red}{\tiny{(-0.117)}}}\\
				
				Strategy3                    
				& {0.916\color{red}{\tiny{(-0.038)}}} &	{0.967\color{red}{\tiny{(-0.007)}}}        & {0.694\color{red}{\tiny{(-0.028)}}} &	{0.881\color{red}{\tiny{(-0.007)}}} & {0.659\color{red}{\tiny{(-0.003)}}} &	{0.892\color{red}{\tiny{(-0.009)}}} \\

				\bottomrule
		\end{tabular}}
		\label{tab_mask}
	\end{table*}
	
	\subsection{Performance Comparison between Different Masking Strategies}  \label{sec_mask}
	In Tab. \ref{tab_mask}, we present the performance comparison between three strategies used in STKIM, WENO, and MHIM-MIL. Specifically, our STKIM masks 10 instances with a probability of 0.6 by default. WENO \cite{qu2022bi} masks 95 instances with a probability of 1.0 by default. MHIM-MIL \cite{tang2023multiple} masks 1\% instances with a probability of 0.5. We find that the masking strategy of our STKIM performs best, significantly suppressing the strategy in WENO and slightly surpassing the strategy in MHIM-MIL.
	
	\subsection{Computational Cost} \label{sec_dis}

	\begin{table}[t]
		\footnotesize
		\centering
		\caption{Comparison of performance and computational cost requirements between MHIM-MIL and STKIM. We report the AUC,  FLOPs,  training time per epoch (Time), and peak memory usage (Mem.) on the CAMELYON-16 (C16) dataset. The flops are measured with the number of instances of a bag being 1024.}
		\vspace{-2mm}
		\begin{tabular}{lcccccc}
			\toprule
			Model           & C16 & BRACS & LBC & FLOPs & Time & Mem.\\ \midrule
			\multicolumn{7}{c}{ResNet18 ImageNet pretrained} \\  \midrule   
			ABMIL & \textbf{0.790} & 0.723 & 0.798 &\textbf{201M} & \textbf{8.0s} & \textbf{0.3G}  \\
			MHIM-MIL & 0.772 & {0.774} & 0.816 &\textbf{201M} &  20.8s & 1.9G \\
			\rowcolor{dino}STKIM& {0.779} & \textbf{0.789} & \textbf{0.820} &\textbf{201M} &  \textbf{8.0s} & \textbf{0.3G} \\\midrule
			\multicolumn{7}{c}{ViT-S/16 SSL pretrained} \\  \midrule   
			ABMIL & 0.945 & {0.866} & 0.831 &\textbf{84M} & \textbf{6.4s} & \textbf{0.2G}  \\
			MHIM-MIL & \textbf{0.970} & {0.865} & \textbf{0.872} &\textbf{84M} &  16.8s & 1.0G \\
			\rowcolor{dino}STKIM& {0.968} & \textbf{0.873} & 0.856 &\textbf{84M} &  6.5s & \textbf{0.2G} \\
			\bottomrule
		\end{tabular}
		
		\label{tab_com_stkim}
	\end{table}	
	
	\begin{table}[t]
		\footnotesize
		\centering
		\caption{Comparison of performance, time, and memory requirements between ABMIL and MBA. We report the auc, the FLOPs, the training time per epoch (Time), and the peak memory usage (Mem.) on the CAMELYON-16 dataset (C16). The flops are measured with the number of instances of a bag being 1024.}
		\label{tab_com_mba}
		\begin{tabular}{lcccccc}
			\toprule
			Model           & C16 & BRACS & LBC & FLOPs & Time & Mem.\\ \midrule
			\multicolumn{7}{c}{ResNet18 ImageNet pretrained} \\  \midrule   
			ABMIL & 0.790 & 0.723 & 0.798 &\textbf{201M} & \textbf{8.0s} & \textbf{0.3G}  \\
			\rowcolor{dino}+MBA& \textbf{0.850} & \textbf{0.797} & \textbf{0.818} &{202M} &  11.6s & \textbf{0.3G} \\ \midrule
			\multicolumn{7}{c}{ViT-S/16 SSL pretrained} \\  \midrule   
			ABMIL & 0.945 & 0.866 & 0.831 &\textbf{84M} & \textbf{6.4s} & \textbf{0.2G}  \\
			\rowcolor{dino}+MBA& \textbf{0.973} & \textbf{0.878} & \textbf{0.875} &{85M} &  9.3s & \textbf{0.2G} \\
			\bottomrule
		\end{tabular}
	\end{table}	
	
	\noindent\textbf{STKIM and MHIM-MIL \cite{tang2023multiple}.} We conducted a comprehensive comparison between STKIM and MHIM-MIL, focusing on computational cost and performance, as detailed in Tab. \ref{tab_com_stkim}. For the computational cost, STKIM demonstrates nearly identical training time consumption and GPU memory usage as the baseline, ABMIL. This similarity arises because STKIM primarily integrates a sorting algorithm, which does not substantially increase resource requirements. On the other hand, MHIM-MIL introduces a teacher model while requiring two forward propagations, leading to significantly higher GPU memory usage and training time consumption. Due to the masking operator being discarded in the evaluation, STKIM and MHIM-MIL keep the same evaluation cost (FLOPs) as ABMIL. For the performance, STKIM delivers comparable results to MHIM-MIL across three datasets and with two pretrained backbone models. Notably, STKIM outperforms MHIM-MIL in four out of six performance metrics while lagging behind in the remaining two.

	\noindent\textbf{MBA.} In Tab. \ref{tab_com_mba}, we present the comparison of performance and computational cost between ABMIL and MBA. Notably, MBA demonstrates a substantial performance improvement over ABMIL. Meanwhile, due to introducing a small number of parameters, the FLOPs, and Memory cost increases marginally. Otherwise, the inclusion of the newly introduced diversity loss leads to a notable increase in time cost.

	\section{Limitations} \label{sec_limitation}
	Although ACMIL enhances the generalization ability and interpretability of MIL methods in WSI analysis, certain limitations necessitate further exploration. Firstly, the selection of hyperparameters $M$ and $K$ significantly impacts performance, and the optimal choice depends on the dataset, requiring practitioners to determine the best value through trial and error. In the future, how to simplify the framework should be considered. Secondly, our paper does not account for the correlation between instances, which is crucial for understanding the complex tumor structure. This aspect will be a focus of future investigations. Thirdly, ACMIL significantly reduces the need for instance annotations compared to instance-supervised approaches and achieves comparable WSI classification performance (AUC: ACMIL 0.974  vs. Full supervised 0.992), but it performs poorly in tumor localization tasks. Tab. \ref{tab:FROC} shows ACMIL achieves an FROC score of 0.4322 on the Camelyon16 tumor localization task, lower than the top-performing supervised approach with a score of 0.8074.

\end{document}